# Reformulating the Situation Calculus and the Event Calculus in the General Theory of Stable Models and in Answer Set Programming

**Joohyung Lee** JOOLEE@ASU.EDU
**Ravi Palla** RAVI.PALLA@ASU.EDU
*School of Computing, Informatics,*
*and Decision Systems Engineering*
*Arizona State University*
*Tempe, AZ 85287, USA*

## Abstract

Circumscription and logic programs under the stable model semantics are two well-known nonmonotonic formalisms. The former has served as a basis of classical logic based action formalisms, such as the situation calculus, the event calculus and temporal action logics; the latter has served as a basis of a family of action languages, such as language $\mathcal{A}$ and several of its descendants. Based on the discovery that circumscription and the stable model semantics coincide on a class of *canonical* formulas, we reformulate the situation calculus and the event calculus in the general theory of stable models. We also present a translation that turns the reformulations further into answer set programs, so that efficient answer set solvers can be applied to compute the situation calculus and the event calculus.

## 1. Introduction

Circumscription (McCarthy, 1980, 1986) and logic programs under the stable model semantics (Gelfond & Lifschitz, 1988) are two well-known nonmonotonic formalisms. As one of the oldest nonmonotonic formalisms, circumscription has found many applications in commonsense reasoning and model-based diagnoses (e.g., McCarthy, 1986; Shanahan, 1995; Besnard & Cordier, 1994). The stable model semantics is the mathematical basis of Answer Set Programming (ASP) (Marek & Truszczyński, 1999; Niemelä, 1999; Lifschitz, 2008), which is being widely applied thanks to the availability of several efficient implementations, known as answer set solvers.

While the two nonmonotonic formalisms have been applied to overlapping classes of problems, minimal model reasoning ensured by circumscription does not coincide with stable model reasoning. Moreover, these formalisms have different roots. While circumscription is defined in terms of translation into *classical (second-order) logic*, stable models proposed by Gelfond and Lifschitz (1988) are defined in terms of *grounding* and *fixpoints* in the style of Reiter's default logic (Reiter, 1980). These differences in part account for the fact that the two formalisms have formed rather disparate traditions in knowledge representation research. In particular, in the area of temporal reasoning, the former has served as a basis of classical logic based action calculi, such as the situation calculus (McCarthy & Hayes, 1969; Reiter, 2001), the event calculus (Shanahan, 1995) and temporal action logics (Doherty,

Gustafsson, Karlsson, & Kvarnström, 1998), whereas the latter has served as a basis of a family of action languages, such as language $\mathcal{A}$ (Gelfond & Lifschitz, 1998) and several of its descendants which can be translated into logic programs under the stable model semantics.

However, a recent generalization of the stable model semantics shed new light on the relationship between circumscription and stable models. The first-order stable model semantics defined by Ferraris, Lee and Lifschitz (2007, 2011) characterizes the stable models of a first-order sentence as the models (in the sense of first-order logic) of the sentence that satisfy the "stability" condition, expressed by a second-order formula that is similar to the one used to define circumscription. Since logic programs are viewed as a special class of first-order sentences under the stable model semantics, this definition extends the stable model semantics by Gelfond and Lifschitz (1988) to the full first-order level without limiting attention to Herbrand models. Essentially the same characterization was independently given by Lin and Zhou (2011), via logic of knowledge and justified assumption (Lin & Shoham, 1992). These definitions are also equivalent to the definition of Quantified Equilibrium Logic given by Pearce and Valverde (2005), which is defined in terms of the logic of Here-and-There (Heyting, 1930).

The new definition of a stable model motivates us to investigate the relationship between stable model reasoning and minimal model reasoning. In particular, we focus on their relationship in the area of temporal reasoning. We show how the situation calculus and the event calculus can be reformulated in the first-order stable model semantics, and further in ASP. This is not only theoretically interesting, but also practically useful as it allows us to leverage efficient answer set solvers for computing circumscriptive action theories.

For this, we develop two technical results. First, we show that circumscription and the first-order stable model semantics coincide on the class of *canonical formulas.* This is the largest syntactic class identified so far on which the two semantics coincide, and is general enough to cover several circumscriptive action formalisms, such as the situation calculus, the event calculus, and temporal action logics. The result allows us to reformulate those action formalisms in the first-order stable model semantics. While minimal model reasoning sometimes leads to unintuitive results, those circumscriptive action formalisms are carefully designed to avoid such cases, and our result implies that minimal model reasoning in those action formalisms can also be viewed as stable model reasoning.

Second, we identify a class of *almost universal* formulas, which can be turned into the syntax of a logic program while preserving stable models. It turns out that the reformulations of the situation calculus and the event calculus in the first-order stable model semantics fall into this class of formulas. We introduce system F2LP that turns formulas in this class to logic programs, and, in conjunction with the result on canonical formulas, use the combination of F2LP and answer set solvers to compute the situation calculus and the event calculus.

Our work makes explicit the relationship between classical logic and logic program traditions in temporal reasoning. Interestingly, the development of the event calculus has spanned over both traditions. The original version of the event calculus (Kowalski & Sergot, 1986) was formulated in logic programs, but not under the stable model semantics (that was the time before the invention of the stable model semantics). More extensive developments have been later carried out on the classical logic foundation via circumscription (e.g., Shanahan, 1995, 1997, 1999; Miller & Shanahan, 1999; Mueller, 2004), but the relation to the logic program formulation remained implicit. Based on the reduction of circumscription to completion, SAT-based event calculus systems were implemented, one by Shanahan and Witkowski (2004) and another by Mueller (2004). The latter system is called the DEC reasoner,[1] which outperforms the former thanks to a more efficient and general compilation method into propositional logic. While the system handles a large fragment of the event calculus, it still cannot handle recursive and disjunctive axioms since completion cannot be applied to such axioms. Our ASP-based approach on the other hand can handle the *full* version of the event calculus under the assumption that the domain is given and finite. Thanks to the efficiency of ASP solvers, our experiments indicate that the ASP-based event calculus reasoner is significantly faster than the DEC reasoner (Appendix B).

Similar to the logic programming tradition of the event calculus, the situation calculus (McCarthy & Hayes, 1969; Reiter, 2001) can be implemented in Prolog, based on the fact that Clark's completion semantics accounts for definitional axioms. But unlike the event calculus, to the best of our knowledge, efficient propositional solvers have not been applied to directly compute the models of situation calculus theories. In this paper, we reformulate Lin's causal action theories (1995) and Reiter's basic action theories (2001) in the first-order stable model semantics and in ASP. For basic action theories, we also provide an ASP-based encoding method that obtains Reiter's successor state axioms from the effect axioms and the *generic* inertia axioms adopted in ASP, the idea of which is close to Reiter's frame default (1980).

The paper is organized as follows. The next section reviews the definitions of circumscription and the first-order stable model semantics, and presents the definition of a canonical formula. Based on this, Sections 3 and 4 reformulate the event calculus and the situation calculus in the first-order stable model semantics. Section 5 shows a translation that turns almost universal formulas into logic programs that can be accepted by ASP solvers. Sections 6 and 7 use this result to turn the reformulations of the event calculus and the situation calculus given in Sections 3 and 4 into the input language of ASP solvers. Complete proofs are given in Appendix C.

## 2. Circumscription and First-Order Stable Model Semantics

We assume the following set of primitive propositional connectives and quantifiers:

$$\bot \text{ (falsity)},\ \wedge,\ \vee,\ \rightarrow,\ \forall,\ \exists\ .$$

We understand $\neg F$ as an abbreviation of $F \rightarrow \bot$; symbol $\top$ stands for $\bot \rightarrow \bot$, and $F \leftrightarrow G$ stands for $(F \rightarrow G) \wedge (G \rightarrow F)$.

### 2.1 Review: Circumscription

Let $\mathbf{p}$ be a list of distinct predicate constants $p_1, \dots, p_n$, and let $\mathbf{u}$ be a list of distinct predicate variables $u_1, \dots, u_n$. By $\mathbf{u} \leq \mathbf{p}$ we denote the conjunction of the formulas $\forall \mathbf{x}(u_i(\mathbf{x}) \rightarrow p_i(\mathbf{x}))$ for all $i = 1, \dots n$, where $\mathbf{x}$ is a list of distinct object variables whose length is the same as the arity of $p_i$. Expression $\mathbf{u} < \mathbf{p}$ stands for $(\mathbf{u} \leq \mathbf{p}) \wedge \neg(\mathbf{p} \leq \mathbf{u})$. For

1. `http://decreasoner.sourceforge.net`

instance, if $p$ and $q$ are unary predicate constants then $(u, v) < (p, q)$ is

$$\forall x(u(x) \to p(x)) \land \forall x(v(x) \to q(x)) \land \neg\Big(\forall x(p(x) \to u(x)) \land \forall x(q(x) \to v(x))\Big).$$

Circumscription is defined in terms of the CIRC operator with *minimized* predicates. For any first-order formula $F$, expression CIRC$[F; \mathbf{p}]$ stands for the second-order formula

$$F \land \neg\exists \mathbf{u}((\mathbf{u} < \mathbf{p}) \land F(\mathbf{u})),$$

where $F(\mathbf{u})$ is the formula obtained from $F$ by substituting $u_i$ for $p_i$. When $F$ is a sentence (i.e., a formula with no free variables), intuitively, the models of CIRC$[F; \mathbf{p}]$ are the models of $F$ that are "minimal" on $\mathbf{p}$.

The definition is straightforwardly extended to the case when $F$ is a many-sorted first-order formula (Lifschitz, 1994, Section 2.4), which is the language that the event calculus and the situation calculus are based on.

## 2.2 Review: First-Order Stable Model Semantics

This review follows the definition by Ferraris et al. (2011). There, the stable models are defined in terms of the SM operator, whose definition is similar to the CIRC operator in the previous section. For any first-order formula $F$ and any finite list of predicate constants $\mathbf{p} = (p_1, \dots, p_n)$, formula SM$[F; \mathbf{p}]$ is defined as

$$F \land \neg\exists \mathbf{u}((\mathbf{u} < \mathbf{p}) \land F^*(\mathbf{u})),$$

where $\mathbf{u}$ is defined the same as in CIRC$[F; \mathbf{p}]$, and $F^*(\mathbf{u})$ is defined recursively as follows:

- $p_i(\mathbf{t})^* = u_i(\mathbf{t})$ for any list $\mathbf{t}$ of terms;
- $F^* = F$ for any atomic formula $F$ (including $\bot$ and equality) that does not contain members of $\mathbf{p}$;
- $(F \land G)^* = F^* \land G^*$;
- $(F \lor G)^* = F^* \lor G^*$;
- $(F \to G)^* = (F^* \to G^*) \land (F \to G)$;
- $(\forall x F)^* = \forall x F^*$;
- $(\exists x F)^* = \exists x F^*$.

The predicates in $\mathbf{p}$ are called *intensional*: these are the predicates that we "intend to characterize" by $F$ in terms of non-intensional predicates.[2] When $F$ is a sentence, the models of the second-order sentence SM$[F; \mathbf{p}]$ are called the $\mathbf{p}$-*stable* models of $F$: they are the models of $F$ that are "stable" on $\mathbf{p}$. We will often simply write SM$[F]$ in place of SM$[F; \mathbf{p}]$ when $\mathbf{p}$ is the list of all predicate constants occurring in $F$. According to Lee, Lifschitz, and

2. Intensional predicates are analogous to output predicates in Datalog, and non-intensional predicates are analogous to input predicates in Datalog (Lifschitz, 2011).

Palla (2008), *answer sets* are defined as a special class of stable models as follows. By $\sigma(F)$ we denote the signature consisting of the object, function and predicate constants occurring in $F$. If $F$ contains at least one object constant, an Herbrand interpretation of $\sigma(F)$ that satisfies SM$[F]$ is called an *answer set* of $F$. The answer sets of a logic program $\Pi$ are defined as the answer sets of the FOL-representation of $\Pi$ (i.e., the conjunction of the universal closures of implications corresponding to the rules). For example, the FOL-representation of the program

$$
\begin{array}{l}
p(a)\\
q(b)\\
r(x) \leftarrow p(x), \mathit{not}\ q(x)
\end{array}
$$

is

$$p(a) \wedge q(b) \wedge \forall x(p(x) \wedge \neg q(x) \rightarrow r(x)) \tag{1}$$

and SM$[F]$ is

$$
\begin{array}{l}
p(a) \wedge q(b) \wedge \forall x(p(x) \wedge \neg q(x) \rightarrow r(x))\\
\quad \wedge \neg \exists uvw(((u,v,w) < (p,q,r)) \wedge u(a) \wedge v(b)\\
\qquad\qquad \wedge \forall x((u(x) \wedge (\neg v(x) \wedge \neg q(x)) \rightarrow w(x)) \wedge (p(x) \wedge \neg q(x) \rightarrow r(x)))),
\end{array}
$$

which is equivalent to the first-order sentence

$$\forall x(p(x) \leftrightarrow x = a) \wedge \forall x(q(x) \leftrightarrow x = b) \wedge \forall x(r(x) \leftrightarrow (p(x) \wedge \neg q(x))) \tag{2}$$

(Ferraris et al., 2007, Example 3). The stable models of $F$ are any first-order models of (2). The only answer set of $F$ is the Herbrand model $\{p(a),\ q(b),\ r(a)\}$.

According to Ferraris et al. (2011), this definition of an answer set, when applied to the syntax of logic programs, is equivalent to the traditional definition of an answer set that is based on grounding and fixpoints (Gelfond & Lifschitz, 1988).

Note that the definition of a stable model is more general than the definition of an answer set in the following ways: stable models are not restricted to Herbrand models, the underlying signature can be arbitrary, and the intensional predicates are not fixed to the list of predicate constants occurring in the formula. The last fact is not essential in view of the following proposition. By $pr(F)$ we denote the list of all predicate constants occurring in $F$; by $\mathit{Choice}(\mathbf{p})$ we denote the conjunction of "choice formulas" $\forall \mathbf{x}(p(\mathbf{x}) \vee \neg p(\mathbf{x}))$ for all predicate constants $p$ in $\mathbf{p}$, where $\mathbf{x}$ is a list of distinct object variables; by $\mathit{False}(\mathbf{p})$ we denote the conjunction of $\forall \mathbf{x} \neg p(\mathbf{x})$ for all predicate constants $p$ in $\mathbf{p}$. We sometimes identify a list with the corresponding set when there is no confusion.

**Proposition 1** *Formula*

$$\text{SM}[F; \mathbf{p}] \leftrightarrow \text{SM}[F \wedge \mathit{Choice}(pr(F) \setminus \mathbf{p}) \wedge \mathit{False}(\mathbf{p} \setminus pr(F))] \tag{3}$$

*is logically valid.*

Notice that the (implicit) intensional predicates on the right-hand side of (3) are those in $(pr(F) \cup \mathbf{p})$. The *Choice* formula makes the predicates in $(pr(F) \setminus \mathbf{p})$ to be exempt from the stability checking. On the other hand, the *False* formula makes the predicates in $(\mathbf{p} \setminus pr(F))$ to be stabilized (i.e., to have empty extents), though they do not occur in $F$.

Ferraris et al. (2011) incorporate strong negation into the stable model semantics by distinguishing between intensional predicates of two kinds, *positive* and *negative*. Each negative intensional predicate has the form $\sim\!p$, where $p$ is a positive intensional predicate and '$\sim$' is a symbol for strong negation. Syntactically $\sim$ is not a logical connective, as it can appear only as a part of a predicate constant. An interpretation of the underlying signature is *coherent* if it satisfies the formula

$$\neg\exists\mathbf{x}(p(\mathbf{x}) \wedge \sim\!p(\mathbf{x})), \tag{4}$$

where $\mathbf{x}$ is a list of distinct object variables, for each negative predicate $\sim\!p$. We usually consider coherent interpretations only. Intuitively, $\sim\!p(\mathbf{t})$ represents that $p(\mathbf{t})$ is false. This is different from $\neg p(\mathbf{t})$ which represents that it is not known that $p(\mathbf{t})$ is true. Similarly, $\neg\sim\!p(\mathbf{t})$ represents that it is not known that $p(\mathbf{t})$ is false, and $\neg\neg p(\mathbf{t})$ represents that it is not known that $p(\mathbf{t})$ is not known to be true. Note that, unlike in first-order logic, $\neg\neg p(\mathbf{t})$ is different from $p(\mathbf{t})$. For instance, formula $p(a)$ has only one answer set $\{p(a)\}$ but $\neg\neg p(a)$ has no answer sets.

Like the extension of circumscription to many-sorted first-order sentences, the definition of a stable model is straightforwardly extended to many-sorted first-order sentences.

### 2.3 Equivalence of the Stable Model Semantics and Circumscription on Canonical Formulas

Neither the stable model semantics nor circumscription is stronger than the other. For example,

$$\mathrm{CIRC}[\forall x(p(x) \vee \neg p(x));\ p] \tag{5}$$

is equivalent to $\forall x \neg p(x)$, and

$$\mathrm{SM}[\forall x(p(x) \vee \neg p(x));\ p] \tag{6}$$

is equivalent to $\top$, so that (5) is stronger than (6). On the other hand,

$$\mathrm{CIRC}[\forall x(\neg p(x) \rightarrow q(x));\ p, q] \tag{7}$$

is equivalent to $\forall x(\neg p(x) \leftrightarrow q(x))$, and

$$\mathrm{SM}[\forall x(\neg p(x) \rightarrow q(x));\ p, q] \tag{8}$$

is equivalent to $\forall x(\neg p(x) \wedge q(x))$, so that (8) is stronger than (7).

In this section, we show that the two semantics coincide on a class of formulas called *canonical formulas*, which we define below. We first review the notions of *positive*, *negative*, and *strictly positive* occurrences.

**Definition 1** *We say that an occurrence of a predicate constant, or any other subexpression, in a formula $F$ is* positive *if the number of implications containing that occurrence in the antecedent is even, and* negative *otherwise. (Recall that we treat $\neg G$ as shorthand for $G \rightarrow \bot$.) We say that the occurrence is* strictly positive *if the number of implications in $F$ containing that occurrence in the antecedent is* 0.

For example, in (1), both occurrences of $q$ are positive, but only the first one is strictly positive.

**Definition 2** *We say that a formula $F$ is* canonical *relative to a list* $\mathbf{p}$ *of predicate constants if*

- *no occurrence of a predicate constant from* $\mathbf{p}$ *is in the antecedents of more than one implication in $F$, and*
- *every occurrence of a predicate constant from* $\mathbf{p}$ *that is in the scope of a strictly positive occurrence of $\exists$ or $\vee$ in $F$ is strictly positive in $F$.*

**Example 1** *The formula*

$$\forall x(\neg p(x) \rightarrow q(x)) \tag{9}$$

*that is shown above is not canonical relative to $\{p, q\}$ since it does not satisfy the first clause of the definition ($p$ occurs in the antecedents of two implications as $\neg p(x)$ is shorthand for $p(x) \rightarrow \bot$). On the other hand, the formula is canonical relative to $\{q\}$. The formula*

$$\forall x(p(x) \vee \neg p(x)) \tag{10}$$

*is not canonical relative to $\{p\}$ since it does not satisfy the second clause (the second occurrence of $p$ is in the scope of a strictly positive occurrence of $\vee$, but is not strictly positive in (10)); the formula*

$$p(a) \wedge (\exists x\, p(x) \rightarrow \exists x\, q(x)) \tag{11}$$

*is canonical relative to $\{p, q\}$, while*

$$p(a, a) \wedge \exists x(p(x, a) \rightarrow p(b, x)) \tag{12}$$

*is not canonical relative to $\{p, q\}$ since it does not satisfy the second clause (the second occurrence of $p$ is in the scope of a strictly positive occurrence of $\exists$, but is not strictly positive in formula (12)).*

The following theorem states that, for any canonical formula, circumscription coincides with the stable model semantics.

**Theorem 1** *For any canonical formula $F$ relative to* $\mathbf{p}$*,*

$$\text{CIRC}[F; \mathbf{p}] \leftrightarrow \text{SM}[F; \mathbf{p}] \tag{13}$$

*is logically valid.*

For instance, for formula (11), which is canonical relative to $\{p, q\}$, formulas CIRC[(11); $p, q$] and SM[(11); $p, q$] are equivalent to each other. Also, any sentence $F$ is clearly canonical relative to $\emptyset$, so that CIRC[$F; \emptyset$] is equivalent to SM[$F; \emptyset$], which in turn is equivalent to $F$. On the other hand, such equivalence may not necessarily hold for non-canonical formulas. For instance, we observed that, for formula (10) that is not canonical relative to $\{p\}$, formulas (5) and (6) are not equivalent to each other. For formula (9) that is not canonical relative to $\{p, q\}$, formulas (7) and (8) are not equivalent to each other. We also observe that formula (12) that is not canonical relative to $\{p, q\}$, CIRC[(12); $p, q$] is not equivalent to SM[(12); $p, q$]: the Herbrand interpretation $\{p(a, a), p(b, a)\}$ satisfies SM[(12); $p, q$], but does not satisfy CIRC[(12); $p, q$].

Note that non-canonical formulas can often be equivalently rewritten as canonical formulas. Since any equivalent transformation preserves the models of circumscription, Theorem 1 can be applied to such non-canonical formulas, by first rewriting them as canonical formulas. For example, formula (9) is equivalent to

$$\forall x(p(x) \lor q(x)), \tag{14}$$

which is canonical relative to $\{p, q\}$, so that CIRC[(9); $p, q$] is equivalent to SM[(14); $p, q$]. For another example, formula (10) is equivalent to

$$\forall x(p(x) \rightarrow p(x)), \tag{15}$$

which is canonical relative to $\{p\}$, so that CIRC[(10); $p$] is equivalent to SM[(15); $p$]. It is clear that this treatment can be applied to any quantifier-free formula (including any propositional formula) because a quantifier-free formula can be equivalently rewritten as a canonical formula by first rewriting it into a clausal normal form and then turning each clause into the form $C \rightarrow D$, where $C$ is a conjunction of atoms and $D$ is a disjunction of atoms.[3]

Sections 3 and 4 use Theorem 1 to reformulate the event calculus and the situation calculus in the first-order stable model semantics.

## 3. Reformulating the Event Calculus in the General Theory of Stable Models

In this section, we review the syntax of circumscriptive event calculus described in Chapter 2 of the book by Mueller (2006). Based on the observation that the syntax conforms to the condition of canonicality, we present a few reformulations of the event calculus in the general theory of stable models.

### 3.1 Review: Circumscriptive Event Calculus

We assume a many-sorted first-order language, which contains an *event* sort, a *fluent* sort, and a *timepoint* sort. A *fluent term* is a term whose sort is a fluent; an *event term* and a *timepoint term* are defined similarly.

**Definition 3** *A* condition *is defined recursively as follows:*

- *If $\tau_1$ and $\tau_2$ are terms, then comparisons $\tau_1 < \tau_2$, $\tau_1 \leq \tau_2$, $\tau_1 \geq \tau_2$, $\tau_1 > \tau_2$, $\tau_1 = \tau_2$, $\tau_1 \neq \tau_2$ are conditions;*

3. It appears unlikely that knowledge has to be encoded in a non-canonical formula such as (12) that cannot be easily turned into an equivalent canonical formula. c.f. "Guide to Axiomatizing Domains in First-Order Logic" (`http://cs.nyu.edu/faculty/davise/guide.html`). It is not a surprise that all circumscriptive action theories mentioned in this paper satisfy the canonicality assumption.

- *If $f$ is a fluent term and $t$ is a timepoint term, then $HoldsAt(f,t)$ and $\neg HoldsAt(f,t)$ are conditions;*
- *If $\gamma_1$ and $\gamma_2$ are conditions, then $\gamma_1 \wedge \gamma_2$ and $\gamma_1 \vee \gamma_2$ are conditions;*
- *If $v$ is a variable and $\gamma$ is a condition, then $\exists v \gamma$ is a condition.*

We will use $e$ and $e_i$ to denote event terms, $f$ and $f_i$ to denote fluent terms, $t$ and $t_i$ to denote timepoint terms, and $\gamma$ and $\gamma_i$ to denote conditions.

In the event calculus, we circumscribe *Initiates*, *Terminates*, and *Releases* to minimize unexpected effects of events, circumscribe *Happens* to minimize unexpected events, and circumscribe $Ab_i$ (abnormality predicates) to minimize abnormalities. Formally, an *event calculus description* is a circumscriptive theory defined as

$$\begin{aligned}&\text{CIRC}[\Sigma\ ;\ \mathit{Initiates}, \mathit{Terminates}, \mathit{Releases}] \wedge \text{CIRC}[\Delta\ ;\ \mathit{Happens}] \\ &\qquad \wedge \text{CIRC}[\Theta\ ;\ Ab_1, \dots, Ab_n] \wedge\ \Xi,\end{aligned} \tag{16}$$

where

- $\Sigma$ is a conjunction of universal closures of axioms of the form

  $$\begin{array}{l}\gamma \rightarrow \mathit{Initiates}(e,f,t)\\ \gamma \rightarrow \mathit{Terminates}(e,f,t)\\ \gamma \rightarrow \mathit{Releases}(e,f,t)\\ \gamma \wedge \pi_1(e,f_1,t) \rightarrow \pi_2(e,f_2,t) \qquad \text{(“effect constraint”)}\\ \gamma \wedge [\neg]\mathit{Happens}(e_1,t) \wedge \cdots \wedge [\neg]\mathit{Happens}(e_n,t) \rightarrow \mathit{Initiates}(e,f,t)\\ \gamma \wedge [\neg]\mathit{Happens}(e_1,t) \wedge \cdots \wedge [\neg]\mathit{Happens}(e_n,t) \rightarrow \mathit{Terminates}(e,f,t),\end{array}$$

  where each of $\pi_1$ and $\pi_2$ is either *Initiates* or *Terminates* (‘$[\neg]$’ means that ‘$\neg$’ is optional);

- $\Delta$ is a conjunction of universal closures of *temporal ordering formulas* (comparisons between timepoint terms) and axioms of the form

  $$\begin{array}{l}\gamma \rightarrow \mathit{Happens}(e,t)\\ \sigma(f,t) \wedge \pi_1(f_1,t) \wedge \cdots \wedge \pi_n(f_n,t) \rightarrow \mathit{Happens}(e,t) \qquad \text{(“causal constraints”)}\\ \mathit{Happens}(e,t) \rightarrow \mathit{Happens}(e_1,t) \vee \cdots \vee \mathit{Happens}(e_n,t) \quad \text{(“disjunctive event axiom”)},\end{array}$$

  where $\sigma$ is *Started* or *Stopped* and each $\pi_j$ $(1 \leq j \leq n)$ is either *Initiated* or *Terminated*;

- $\Theta$ is a conjunction of universal closures of *cancellation* axioms of the form

  $$\gamma \rightarrow Ab_i(...,t)\ ;$$

- $\Xi$ is a conjunction of first-order sentences (outside the scope of CIRC) including unique name axioms, state constraints, event occurrence constraints, and the set of domain-independent axioms in the event calculus, such as *EC* (for the continuous event calculus) and *DEC* (for the discrete event calculus) (Mueller, 2006, Chapter 2). It also includes the following definitions of the predicates used in the causal constraints in $\Delta$:

$$\mathit{Started}(f,t) \overset{def}{\leftrightarrow} (\mathit{HoldsAt}(f,t) \lor \exists e(\mathit{Happens}(e,t) \land \mathit{Initiates}(e,f,t))) \quad (CC_1)$$
$$\mathit{Stopped}(f,t) \overset{def}{\leftrightarrow} (\neg \mathit{HoldsAt}(f,t) \lor \exists e(\mathit{Happens}(e,t) \land \mathit{Terminates}(e,f,t))) \quad (CC_2)$$
$$\mathit{Initiated}(f,t) \overset{def}{\leftrightarrow} (\mathit{Started}(f,t) \land \neg\exists e(\mathit{Happens}(e,t) \land \mathit{Terminates}(e,f,t))) \quad (CC_3)$$
$$\mathit{Terminated}(f,t) \overset{def}{\leftrightarrow} (\mathit{Stopped}(f,t) \land \neg\exists e(\mathit{Happens}(e,t) \land \mathit{Initiates}(e,f,t))) \quad (CC_4).$$

**Remark 1** *The following facts are easy to check:*

- $\Sigma$ *is canonical relative to* $\{\mathit{Initiates}, \mathit{Terminates}, \mathit{Releases}\}$*;*
- $\Delta$ *is canonical relative to* $\{\mathit{Happens}\}$*;*
- $\Theta$ *is canonical relative to* $\{Ab_1, \dots, Ab_n\}$*.*

These facts are used in the next section to reformulate the event calculus in the general theory of stable models.

### 3.2 Reformulating the Event Calculus in the General Theory of Stable Models

Following Ferraris, Lee, Lifschitz, and Palla (2009), about a formula $F$ we say that it is *negative* on a list $\mathbf{p}$ of predicate constants if members of $\mathbf{p}$ have no strictly positive occurrences in $F$.[4] For example, formula (9) is negative on $\{p\}$, but is not negative on $\{p,q\}$. A formula of the form $\neg F$ (shorthand for $F \to \bot$) is negative on any list of predicates.

We assume that $\Xi$ was already equivalently rewritten so that $\Xi$ is negative on $\{\mathit{Initiates}$, $\mathit{Terminates}$, $\mathit{Releases}$, $\mathit{Happens}$, $Ab_1, \dots, Ab_n\}$. This can be easily done by prepending $\neg\neg$ to strictly positive occurrences of those predicates. The following theorem shows a few equivalent reformulations of circumscriptive event calculus in the general theory of stable models.

**Theorem 2** *For any event calculus description (16), the following theories are equivalent to each other:*[5]

*(a)* $\mathrm{CIRC}[\Sigma; I, T, R] \land \mathrm{CIRC}[\Delta; H] \land \mathrm{CIRC}[\Theta; Ab_1, \dots, Ab_n] \land \Xi$ ;

*(b)* $\mathrm{SM}[\Sigma; I, T, R] \land \mathrm{SM}[\Delta; H] \land \mathrm{SM}[\Theta; Ab_1, \dots, Ab_n] \land \Xi$ ;

*(c)* $\mathrm{SM}[\Sigma \land \Delta \land \Theta \land \Xi; I, T, R, H, Ab_1, \dots, Ab_n]$ ;

*(d)* $\mathrm{SM}[\Sigma \land \Delta \land \Theta \land \Xi \land \mathit{Choice}(\mathit{pr}(\Sigma \land \Delta \land \Theta \land \Xi) \setminus \{I, T, R, H, Ab_1, \dots, Ab_n\})]$ .

The equivalence between (a) and (b) is immediate from Theorem 1. The equivalence between (b) and (c) can be shown using the splitting theorem by Ferraris et al. (2009). The assumption that $\Xi$ is negative on the intensional predicates is essential in showing that

4. Note that we distinguish between a formula being negative (on $\mathbf{p}$) and an occurrence being negative (Section 2.3).
5. For brevity, we abbreviate the names of circumscribed predicates.

equivalence (For more details, see the proof in Appendix C.4.). The equivalence between (c) and (d) follows from Proposition 1 since

$$\{I, T, R, H, Ab_1, \dots, Ab_n\} \setminus pr(\Sigma \wedge \Delta \wedge \Theta \wedge \Xi)$$

is the empty set.[6]

## 4. Reformulating the Situation Calculus in the General Theory of Stable Models

In this section, we review and reformulate two versions of the situation calculus—Lin's causal action theories (1995) and Reiter's basic action theories (2001).

### 4.1 Review: Lin's Causal Action Theories

We assume a many-sorted first-order language which contains a *situation* sort, an *action* sort, a *fluent* sort, a *truth value* sort and an *object* sort. We understand expression $P(\mathbf{x}, s)$, where $P$ is a fluent name, as shorthand for $Holds(P(\mathbf{x}), s)$. We do not consider functional fluents here for simplicity.

According to Lin (1995), a formula $\phi(s)$ is called a *simple state formula about s* if $\phi(s)$ does not mention *Poss*, *Caused* or any situation term other than possibly the variable $s$.

We assume that a causal action theory $\mathcal{D}$ consists of a finite number of the following sets of axioms. We often identify $\mathcal{D}$ with the conjunction of the universal closures of all axioms in $\mathcal{D}$. In the following, $F$, $F_i$ are fluent names, $A$ is an action name, $V$, $V_i$ are truth values, $s$, $s'$ are situation variables, $\phi(s)$ is a simple state formula about $s$, symbols $a$, $a'$ are action variables, $f$ is a variable of sort fluent, $v$ is a variable of sort truth value, and $\mathbf{x}$, $\mathbf{x}_i$, $\mathbf{y}$, $\mathbf{y}_i$ are lists of variables.

- $\mathcal{D}_{caused}$ is a conjunction of axioms of the form

  $$Poss(A(\mathbf{x}), s) \rightarrow (\phi(s) \rightarrow Caused(F(\mathbf{y}), V, do(A(\mathbf{x}), s))$$

  (direct effect axioms), and

  $$\phi(s) \wedge Caused(F_1(\mathbf{x}_1), V_1, s) \wedge \cdots \wedge Caused(F_n(\mathbf{x}_n), V_n, s) \rightarrow Caused(F(\mathbf{x}), V, s)$$

  (indirect effect axioms).
- $\mathcal{D}_{poss}$ is a conjunction of precondition axioms of the form

  $$Poss(A(\mathbf{x}), s) \leftrightarrow \phi(s). \tag{17}$$

- $\mathcal{D}_{rest}$ is a conjunction of the following axioms:
  - The basic axioms:

    $$Caused(f, true, s) \rightarrow Holds(f, s),$$
    $$Caused(f, false, s) \rightarrow \neg Holds(f, s),$$

    $$true \neq false \wedge \forall v(v = true \vee v = false). \tag{18}$$

6. $I$, $T$, $R$, $H$ occur in the domain independent axioms as part of $\Xi$.

– The unique name assumptions for fluent names:

$$\begin{array}{l} F_i(\mathbf{x}) \neq F_j(\mathbf{y}), \quad (i \neq j) \\ F_i(\mathbf{x}) = F_i(\mathbf{y}) \rightarrow \mathbf{x} = \mathbf{y}. \end{array} \tag{19}$$

Similarly for action names.

– The *foundational axioms* for the discrete situation calculus: [7]

$$s \neq do(a, s), \tag{20}$$

$$do(a, s) = do(a', s') \rightarrow (a = a' \wedge s = s'), \tag{21}$$

$$\forall p \Big( p(S_0) \wedge \forall a, s\big(p(s) \rightarrow p(do(a, s))\big) \rightarrow \forall s\ p(s) \Big). \tag{22}$$

– The frame axiom:

$$\begin{array}{l} Poss(a, s) \rightarrow (\neg \exists v\, Caused(f, v, do(a, s)) \\ \qquad\qquad\qquad \rightarrow (Holds(f, do(a, s)) \leftrightarrow Holds(f, s))). \end{array}$$

– Axioms for other domain knowledge: $\phi(s)$.

A *causal action theory* is defined as

$$\mathrm{CIRC}[\mathcal{D}_{caused};\ Caused] \wedge \mathcal{D}_{poss} \wedge \mathcal{D}_{rest}. \tag{23}$$

**Remark 2** *It is easy to check that $\mathcal{D}_{caused}$ is canonical relative to Caused.*

This fact is used in the next section to reformulate causal action theories in the general theory of stable models.

### 4.2 Reformulating Causal Action Theories in the General Theory of Stable Models

Let $\mathcal{D}_{poss^{\rightarrow}}$ be the conjunction of axioms $\phi(s) \rightarrow Poss(A(\mathbf{x}), s)$ for each axiom (17) in $\mathcal{D}_{poss}$. Instead of the second-order axiom (22), we consider the following first-order formula $\mathcal{D}_{sit}$, which introduces a new intensional predicate constant *Sit* whose argument sort is situation.[8]

$$Sit(S_0) \wedge \forall a, s(Sit(s) \rightarrow Sit(do(a, s))) \wedge \neg \exists s \neg Sit(s). \tag{24}$$

In the following, $\mathcal{D}^{-}_{rest}$ is the theory obtained from $\mathcal{D}_{rest}$ by dropping (22).

**Theorem 3** *Given a causal action theory (23), the following theories are equivalent to each other when we disregard the auxiliary predicate Sit:*

*(a)* $\mathrm{CIRC}[\mathcal{D}_{caused}; Caused] \wedge \mathcal{D}_{poss} \wedge \mathcal{D}_{rest}$ *;*

*(b)* $\mathrm{SM}[\mathcal{D}_{caused}; Caused] \wedge \mathcal{D}_{poss} \wedge \mathcal{D}^{-}_{rest} \wedge \mathrm{SM}[\mathcal{D}_{sit}; Sit]$ *;*

*(c)* $\mathrm{SM}[\mathcal{D}_{caused}; Caused] \wedge \mathrm{SM}[\mathcal{D}_{poss^{\rightarrow}};\ Poss] \wedge \mathcal{D}^{-}_{rest} \wedge \mathrm{SM}[\mathcal{D}_{sit}; Sit]$ *;*

*(d)* $\mathrm{SM}[\mathcal{D}_{caused} \wedge \mathcal{D}_{poss^{\rightarrow}} \wedge \mathcal{D}^{-}_{rest} \wedge \mathcal{D}_{sit};\ Caused, Poss, Sit]$ *.*

7. For simplicity we omit two other axioms regarding the partial-order among situations.
8. Suggested by Vladimir Lifschitz (personal communication).

### 4.3 Review: Reiter's Basic Action Theories

As in causal action theories, we understand $P(\mathbf{x}, s)$, where $P$ is a fluent name, as shorthand for $Holds(P(\mathbf{x}), s)$, and do not consider functional fluents.

A *basic action theory (BAT)* is of the form

$$\Sigma \cup \mathcal{D}_{ss} \cup \mathcal{D}_{ap} \cup \mathcal{D}_{una} \cup \mathcal{D}_{S_0}\ , \tag{25}$$

where

- $\Sigma$ is the conjunction of the foundational axioms (Section 4.1);
- $\mathcal{D}_{ss}$ is a conjunction of successor state axioms of the form

  $$F(\mathbf{x}, do(a, s)) \leftrightarrow \Phi_F(\mathbf{x}, a, s),$$

  where $\Phi_F(\mathbf{x}, a, s)$ is a formula that is uniform in $s$ [9] and whose free variables are among $\mathbf{x}, a, s$;
- $\mathcal{D}_{ap}$ is a conjunction of action precondition axioms of the form

  $$Poss(A(\mathbf{x}), s) \leftrightarrow \Pi_A(\mathbf{x}, s),$$

  where $\Pi_A(\mathbf{x}, s)$ is a formula that is uniform in $s$ and whose free variables are among $\mathbf{x}, s$;
- $\mathcal{D}_{una}$ is the conjunction of unique name axioms for fluents and actions;
- $\mathcal{D}_{S_0}$ is a conjunction of first-order formulas that are uniform in $S_0$.

### 4.4 Reformulating Basic Action Theories in the General Theory of Stable Models

Note that a BAT is a theory in first-order logic.[10] In view of the fact that any first-order logic sentence $F$ is equivalent to SM$[F; \emptyset]$, it is trivial to view a BAT as a first-order theory under the stable model semantics with the list of intensional predicates being empty.

In the rest of this section, we consider an alternative encoding of BAT in ASP, in which we do not need to provide explicit successor state axioms $\mathcal{D}_{ss}$. Instead, the successor state axioms are entailed by the effect axioms and the generic inertia axioms adopted in ASP by making intensional both the positive predicate *Holds* and the negative predicate $\sim$*Holds* (Recall the definitions of positive and negative predicates in Section 2.2). In the following we assume that the underlying signature contains both these predicates.

An ASP-style BAT is of the form

$$\Sigma \cup \mathcal{D}_{\mathit{effect}} \cup \mathcal{D}_{\mathit{precond}} \cup \mathcal{D}_{\mathit{inertia}} \cup \mathcal{D}_{\mathit{exogenous_0}} \cup \mathcal{D}_{una} \cup \mathcal{D}_{S_0}, \tag{26}$$

where

- $\Sigma$, $\mathcal{D}_{una}$ and $\mathcal{D}_{S_0}$ are defined as before;

9. We refer the reader to the book by Reiter (2001) for the definition of a uniform formula.
10. For simplicity we disregard the second-order axiom (22).

- $\mathcal{D}_{effect}$ is a conjunction of axioms of the form

$$\gamma_R^+(\mathbf{x}, a, s) \rightarrow Holds(R(\mathbf{x}), do(a, s)) \tag{27}$$

or

$$\gamma_R^-(\mathbf{x}, a, s) \rightarrow \sim\!Holds(R(\mathbf{x}), do(a, s)), \tag{28}$$

where $\gamma_R^+(\mathbf{x}, a, s)$ and $\gamma_R^-(\mathbf{x}, a, s)$ are formulas that are uniform in $s$ and whose free variables are among $\mathbf{x}, a$ and $s$;

- $\mathcal{D}_{precond}$ is a conjunction of axioms of the form

$$\pi_A(\mathbf{x}, s) \rightarrow Poss(A(\mathbf{x}), s), \tag{29}$$

where $\pi_A(\mathbf{x}, s)$ is a formula that is uniform in $s$ and whose free variables are among $\mathbf{x}, s$;

- $\mathcal{D}_{inertia}$ is the conjunction of the axioms

$$Holds(R(\mathbf{x}), s) \wedge \neg \sim\!Holds(R(\mathbf{x}), do(a, s)) \rightarrow \ Holds(R(\mathbf{x}), do(a, s)),$$
$$\sim\!Holds(R(\mathbf{x}), s) \wedge \neg Holds(R(\mathbf{x}), do(a, s)) \rightarrow \sim\!Holds(R(\mathbf{x}), do(a, s))$$

for all fluent names $R$;

- $\mathcal{D}_{exogenous_0}$ is the conjunction of

$$Holds(R(\mathbf{x}), S_0) \vee \sim\!Holds(R(\mathbf{x}), S_0)$$

for all fluent names $R$.

Note that axioms in $\mathcal{D}_{inertia}$ are typically used in answer set programming to represent the common sense law of inertia (Lifschitz & Turner, 1999). Similarly, $\mathcal{D}_{exogenous_0}$ is used to represent that the initial value of a fluent is arbitrary.[11]

We will show how this ASP-style BAT is related to Reiter's BAT. First, since we use strong negation, it is convenient to define the following notions. Given the signature $\sigma$ of a BAT, $\sigma^{Holds}$ is the signature obtained from $\sigma$ by adding $\sim\!Holds$ to $\sigma$. We say that an interpretation $I$ of $\sigma^{Holds}$ is *complete* on *Holds* if it satisfies

$$\forall \mathbf{y}(Holds(\mathbf{y}) \vee \sim\!Holds(\mathbf{y})),$$

where $\mathbf{y}$ is a list of distinct variables. Given an interpretation $I$ of $\sigma^{Holds}$, expression $I|_\sigma$ denotes the projection of $I$ on $\sigma$.

Let $\mathcal{D}_{ss}$ be the conjunction of successor state axioms

$$Holds(R(\mathbf{x}), do(a, s)) \ \leftrightarrow \ \Gamma_R^+(\mathbf{x}, a, s) \vee (Holds(R(\mathbf{x}), s) \wedge \neg\Gamma_R^-(\mathbf{x}, a, s)),$$

where $\Gamma_R^+(\mathbf{x}, a, s)$ is the disjunction of $\gamma_R^+(\mathbf{x}, a, s)$ for all axioms (27) in $\mathcal{D}_{effect}$, and $\Gamma_R^-(\mathbf{x}, a, s)$ is the disjunction of $\gamma_R^-(\mathbf{x}, a, s)$ for all axioms (28) in $\mathcal{D}_{effect}$. By $\mathcal{D}_{ap}$ we denote the conjunction of axioms $Poss(A(\mathbf{x}), s) \leftrightarrow \Pi_A(\mathbf{x}, s)$, where $\Pi_A(\mathbf{x}, s)$ is the disjunction of $\pi_A(\mathbf{x}, s)$ for all axioms (29) in $\mathcal{D}_{precond}$.

11. The axioms $\mathcal{D}_{inertia}$ and $\mathcal{D}_{exogenous_0}$ are also closely related to the translation of $\mathcal{C}+$ into nonmonotonic causal logic (Giunchiglia, Lee, Lifschitz, McCain, & Turner, 2004).

**Theorem 4** *Let $T$ be a theory (26) of signature $\sigma^{Holds}$, and $I$ a coherent interpretation of $\sigma^{Holds}$ that is complete on $Holds$. If $I$ satisfies*

$$\neg\exists \mathbf{x}\, a\, s(\Gamma^+_R(\mathbf{x}, a, s) \wedge \Gamma^-_R(\mathbf{x}, a, s))$$

*for every fluent name $R$, then $I$ satisfies*

$$\mathrm{SM}[T;\ Poss, Holds, \sim\!Holds]$$

*iff $I|_\sigma$ satisfies the BAT*

$$\Sigma \wedge \mathcal{D}_{ss} \wedge \mathcal{D}_{ap} \wedge \mathcal{D}_{una} \wedge \mathcal{D}_{S_0}.$$

## 5. Translating Almost Universal Sentences into Logic Programs

Theorems 2—4 present reformulations of the situation calculus and the event calculus in the general theory of stable models, which may contain nested quantifiers and connectives. On the other hand, the input languages of ASP solvers are limited to simple rule forms, which are analogous to clausal normal form in classical logic. Although any first-order formula can be rewritten in clausal normal form while preserving satisfiability, such transformations do not necessarily preserve stable models. This is due to the fact that the notion of equivalence is "stronger" under the stable model semantics (Lifschitz, Pearce, & Valverde, 2001).

**Definition 4** *(Ferraris et al., 2011) A formula $F$ is strongly equivalent to formula $G$ if, for any formula $H$ containing $F$ as a subformula (and possibly containing object, function and predicate constants that do not occur in $F$, $G$), and for any list $\mathbf{p}$ of distinct predicate constants,* $\mathrm{SM}[H; \mathbf{p}]$ *is equivalent to* $\mathrm{SM}[H'; \mathbf{p}]$*, where $H'$ is obtained from $H$ by replacing an occurrence of $F$ by $G$.*

In other words, replacing a subformula with another strongly equivalent subformula does not change the stable models of the whole formula. While strongly equivalent theories are classically equivalent (i.e., equivalent under classical logic), the converse does not hold. Consequently, classically equivalent transformations do not necessarily preserve stable models. For instance, consider $p$ and $\neg\neg p$. When $p$ is intensional, the former has stable models and the latter does not.

It is known that every propositional formula can be rewritten as a logic program (Cabalar & Ferraris, 2007; Cabalar, Pearce, & Valverde, 2005; Lee & Palla, 2007), and such translations can be extended to quantifier-free formulas in a straightforward way (Section 5.1). However, the method does not work in the presence of arbitrary quantifiers, because in the target formalism (logic programs), all variables are implicitly universally quantified.

In this section, we present a translation that turns a certain class of sentences called "almost universal" sentences into logic programs while preserving stable models. It turns out that the reformulations of the situation calculus and the event calculus in Sections 3 and 4 belong to the class of almost universal sentences, so that we can use ASP solvers for computing them.

### 5.1 Translating Quantifier-Free Formulas into Logic Programs

Cabalar et al. (2005) define the following transformation that turns any propositional formula under the stable model semantics into a logic program.

- Left side rules:

$$\top \wedge F \rightarrow G \quad\mapsto\quad \{F \rightarrow G\} \tag{L1}$$

$$\bot \wedge F \rightarrow G \quad\mapsto\quad \emptyset \tag{L2}$$

$$\neg\neg F \wedge G \rightarrow H \quad\mapsto\quad \{G \rightarrow \neg F \vee H\} \tag{L3}$$

$$(F \vee G) \wedge H \rightarrow K \quad\mapsto\quad \left\{ \begin{array}{l} F \wedge H \rightarrow K \\ G \wedge H \rightarrow K \end{array} \right\} \tag{L4}$$

$$(F \rightarrow G) \wedge H \rightarrow K \quad\mapsto\quad \left\{ \begin{array}{l} \neg F \wedge H \rightarrow K \\ G \wedge H \rightarrow K \\ H \rightarrow F \vee \neg G \vee K \end{array} \right\} \tag{L5}$$

- Right side rules:

$$F \rightarrow \bot \vee G \quad\mapsto\quad \{F \rightarrow G\} \tag{R1}$$

$$F \rightarrow \top \vee G \quad\mapsto\quad \emptyset \tag{R2}$$

$$F \rightarrow \neg\neg G \vee H \quad\mapsto\quad \{\neg G \wedge F \rightarrow H\} \tag{R3}$$

$$F \rightarrow (G \wedge H) \vee K \quad\mapsto\quad \left\{ \begin{array}{l} F \rightarrow G \vee K \\ F \rightarrow H \vee K \end{array} \right\} \tag{R4}$$

$$F \rightarrow (G \rightarrow H) \vee K \quad\mapsto\quad \left\{ \begin{array}{l} G \wedge F \rightarrow H \vee K \\ \neg H \wedge F \rightarrow \neg G \vee K \end{array} \right\} \tag{R5}$$

Before applying this transformation to each formula on the lefthand side, we assume that the formula is already written in *negation normal form*, in which negation is applied to literals only, by using the following transformation:

- Negation normal form conversion:

$$\begin{array}{lcl} \neg\top & \mapsto & \bot \\ \neg\bot & \mapsto & \top \\ \neg\neg\neg F & \mapsto & \neg F \\ \neg(F \wedge G) & \mapsto & \neg F \vee \neg G \\ \neg(F \vee G) & \mapsto & \neg F \wedge \neg G \\ \neg(F \rightarrow G) & \mapsto & \neg\neg F \wedge \neg G \end{array}$$

According to Cabalar et al. (2005), successive application of the rewriting rules above turn any propositional formula into a disjunctive logic program. This result can be simply extended to turn any quantifier-free formula into a logic program.

As noted by Cabalar et al. (2005), this translation may involve an exponential blowup in size, and Theorem 1 from their paper shows that indeed there is no vocabulary-preserving polynomial time algorithm to convert general propositional theories under the stable model semantics into disjunctive logic programs. Alternatively, one can use another translation from the same paper, which is linear in size but involves auxiliary atoms and is more complex.

### 5.2 Quantifier Elimination

We introduce a quantifier elimination method that distinguishes between two kinds of occurrences of quantifiers: "singular" and "non-singular." Any "non-singular" occurrence of a quantifier is easy to eliminate, while a "singular" occurrence is eliminated under a certain syntactic condition.

**Definition 5** *We say that an occurrence of $QxG$ in $F$ is* singular *if*

- *$Q$ is $\exists$, and the occurrence of $QxG$ is positive in $F$, or*
- *$Q$ is $\forall$, and the occurrence of $QxG$ is negative in $F$.*

For example, the occurrence of $\exists x\, q(x)$ is singular in (11), but the occurrence of $\exists x\, p(x)$ is not.

Non-singular occurrences of quantifiers can be eliminated in view of the fact that every first-order sentence can be rewritten in prenex form. The prenex form conversion rules given in Section 6.3.1 of Pearce and Valverde (2005) preserve strong equivalence, which leads to the following theorem.[12]

**Theorem 5** *(Lee & Palla, 2007, Proposition 5) Every first-order formula is strongly equivalent to a formula in prenex form.*

The prenex form conversion turns a non-singular occurrence of a quantifier into an outermost $\forall$ while preserving strong equivalence. Consequently, if a sentence contains no singular occurrence of a quantifier, then the above results can be used to turn the sentence into a universal sentence and then into a set of ASP rules. However, in the presence of a singular occurrence of a quantifier, the prenex form conversion turns the occurrence into an outermost $\exists$, which is not allowed in logic programs. Below we consider how to handle such occurrences.

Obviously, if the Herbrand universe is finite, and if we are interested in Herbrand stable models (i.e., answer sets) only, quantified formulas can be rewritten as multiple disjunctions and conjunctions. We do not even need to consider turning the formula into prenex form. For example, for a formula

$$r \wedge \neg\exists x(p(x) \wedge q(x)) \rightarrow s \tag{30}$$

occurring in a theory whose signature contains $\{1, \ldots, n\}$ as the only object constants (and no other function constants), if we replace $\exists x(p(x) \wedge q(x))$ with multiple disjunctions and then turn the resulting program with nested expressions into a usual disjunctive program (Lifschitz, Tang, & Turner, 1999), $2^n$ rules are generated. For instance, if $n = 3$, the

12. Pearce and Valverde (2005) show that a sentence in $\mathbf{QN}_5^c$, the monotonic basis of Quantified Equilibrium Logic, can be turned into prenex form, from which the result follows.

resulting logic program is

$$\begin{array}{l}
s \leftarrow r, not\ p(1), not\ p(2), not\ p(3)\\
s \leftarrow r, not\ p(1), not\ p(2), not\ q(3)\\
s \leftarrow r, not\ p(1), not\ q(2), not\ p(3)\\
s \leftarrow r, not\ p(1), not\ q(2), not\ q(3)\\
s \leftarrow r, not\ q(1), not\ p(2), not\ p(3)\\
s \leftarrow r, not\ q(1), not\ p(2), not\ q(3)\\
s \leftarrow r, not\ q(1), not\ q(2), not\ p(3)\\
s \leftarrow r, not\ q(1), not\ q(2), not\ q(3).
\end{array}$$

Also, the translation is not modular as it depends on the underlying domain; the multiple disjunctions or conjunctions need to be updated when the domain changes. More importantly, this method is not applicable if the theory contains function constants of positive arity, as its Herbrand universe is infinite.

One may also consider introducing Skolem constants as in first-order logic, presuming that, for any sentence $F$ and its "Skolem form" $F'$, $\mathrm{SM}[F;\mathbf{p}]$ is satisfiable iff $\mathrm{SM}[F';\mathbf{p}]$ is satisfiable. However, this idea does not work.[13]

**Example 2** *For formula*

$$F = (\forall x\, p(x) \rightarrow q) \wedge \neg\neg\exists x(q \wedge \neg p(x)),$$

$\mathrm{SM}[F;q]$ *is equivalent to the first-order sentence*

$$(q \leftrightarrow \forall x\, p(x)) \wedge \exists x(q \wedge \neg p(x)),$$

*which is unsatisfiable (the equivalence can be established using Theorems 3 and 11 from Ferraris et al., 2011). Formula $F$ is strongly equivalent to its prenex form*

$$\exists x \exists y\big((p(x) \rightarrow q) \wedge \neg\neg(q \wedge \neg p(y))\big), \tag{31}$$

*However, if we introduce new object constants $a$ and $b$ to replace the existentially quantified variables as in*

$$F' = (p(a) \rightarrow q) \wedge \neg\neg(q \wedge \neg p(b)),$$

*formula* $\mathrm{SM}[F';q]$ *is equivalent to*

$$(q \leftrightarrow p(a)) \wedge (q \wedge \neg p(b)),$$

*which is satisfiable.*

Here we present a method of eliminating singular occurrences of quantifiers by introducing auxiliary predicates. Our idea is a generalization of the practice in logic programming

13. Pearce and Valverde (2005) show that Skolemization works with $\mathbf{QN}_5^c$, the monotonic basis of Quantified Equilibrium Logic, but as our example shows, this does not imply that Skolemization works with Quantified Equilibrium Logic.

that simulates negated existential quantification in the body of a rule by introducing auxiliary predicates. For instance, in order to eliminate $\exists$ in (30), we will introduce a new predicate constant $p'$, and turn (30) into

$$(r \wedge \neg p' \rightarrow s) \wedge \forall x(p(x) \wedge q(x) \rightarrow p'), \tag{32}$$

which corresponds to the logic program

$$\begin{array}{rcl} s & \leftarrow & r, \mathit{not}\ p' \\ p' & \leftarrow & p(x), q(x). \end{array} \tag{33}$$

The models of SM[(30); $p, q, r, s$] are the same as the stable models of (33) if we disregard $p'$. This method does not involve grounding, so that the translation does not depend on the domain and is not restricted to Herbrand models. The method is formally justified by the following proposition.

Recall that a formula $H$ is *negative on* $\mathbf{p}$ if members of $\mathbf{p}$ have no strictly positive occurrences in $H$. Given a formula $F$, we say that an occurrence of a subformula $G$ is $\mathbf{p}$-*negated* in $F$ if it is contained in a subformula $H$ of $F$ that is negative on $\mathbf{p}$.

**Proposition 2** *Let $F$ be a sentence, let $\mathbf{p}$ be a finite list of distinct predicate constants, and let $q$ be a new predicate constant that does not occur in $F$. Consider any non-strictly positive, $\mathbf{p}$-negated occurrence of $\exists y G(y, \mathbf{x})$ in $F$, where $\mathbf{x}$ is the list of all free variables of $\exists y G(y, \mathbf{x})$. Let $F'$ be the formula obtained from $F$ by replacing that occurrence of $\exists y G(y, \mathbf{x})$ with $q(\mathbf{x})$. Then*

$$\mathrm{SM}[F; \mathbf{p}] \wedge \forall \mathbf{x}(q(\mathbf{x}) \leftrightarrow \exists y G(y, \mathbf{x}))$$

*is equivalent to*

$$\mathrm{SM}[F' \wedge \forall \mathbf{x} y(G(y, \mathbf{x}) \rightarrow q(\mathbf{x}));\ \mathbf{p}, q].$$

Proposition 2 tells us that SM[$F$; $\mathbf{p}$] and SM[$F' \wedge \forall \mathbf{x} y(G(y, \mathbf{x}) \rightarrow q(\mathbf{x})$; $\mathbf{p}, q$] have the same models if we disregard the new predicate constant $q$. Notice that $F'$ does not retain the occurrence of $\exists y$.

**Example 3** *In formula (30), $\exists x(p(x) \wedge q(x))$ is contained in a negative formula (relative to any set of intensional predicates). In accordance with Proposition 2,* SM[(30); $p, q, r, s$] *has the same models as* SM[(32); $p, q, r, s, p'$] *if we disregard $p'$.*

Any singular, $\mathbf{p}$-negated occurrence of a subformula $\forall y G(y, \mathbf{x})$ can also be eliminated using Proposition 2 by first rewriting $\forall y G(y, \mathbf{x})$ as $\neg \exists y \neg G(y, \mathbf{x})$. Note that $\forall y G(y, \mathbf{x})$ is not strongly equivalent to $\neg \exists y \neg G(y, \mathbf{x})$, and in general such a classically equivalent transformation may not necessarily preserve stable models. However, the Theorem on Double Negations (Ferraris et al., 2009, also reviewed in Appendix C) tells us that such a transformation is ensured to preserve $\mathbf{p}$-stable models if the replaced occurrence is $\mathbf{p}$-negated in the given formula.

Now we are ready to present our quantifier elimination method, which applies to the class of *almost universal formulas*.

**Definition 6** *We say that a formula $F$ is almost universal relative to* $\mathbf{p}$ *if every singular occurrence of $QxG$ in $F$ is* $\mathbf{p}$*-negated in $F$.*

For example, formula (30) is almost universal relative to any set of predicates because the only singular occurrence of $\exists x(p(x) \wedge q(x))$ in (30) is contained in $\neg\exists x(p(x) \wedge q(x))$, which is negative on any list of predicates. Formula $F$ in Example 2 is almost universal relative to $\{q\}$ because the singular occurrence of $\forall x\, p(x)$ is contained in the formula itself, which is negative on $\{q\}$, and the singular occurrence of $\exists x(q \wedge \neg p(x))$ is contained in $\neg\exists x(q \wedge \neg p(x))$, which is also negative on $\{q\}$.

The following procedure can be used to eliminate all (possibly nested) quantifiers in any almost universal sentence.

**Definition 7 (Translation ELIM-QUANTIFIERS)** *Given a formula $F$, first prepend $\neg\neg$ to every maximal strictly positive occurrence of a formula of the form $\exists y H(y, \mathbf{x})$,*[14] *and then repeat the following process until there are no occurrences of quantifiers remaining: Select a maximal occurrence of a formula of the form $QyG(y, \mathbf{x})$ in $F$, where $Q$ is $\forall$ or $\exists$, and $\mathbf{x}$ is the list of all free variables in $QyG(y, \mathbf{x})$.*

*(a) If the occurrence of $QyG(y, \mathbf{x})$ in $F$ is non-singular in $F$, then set $F$ to be the formula obtained from $F$ by replacing the occurrence of $QyG(y, \mathbf{x})$ with $G(z, \mathbf{x})$, where $z$ is a new variable.*

*(b) Otherwise, if $Q$ is $\exists$ and the occurrence of $QyG(y, \mathbf{x})$ in $F$ is positive, then set $F$ to be*

$$F' \wedge (G(y, \mathbf{x}) \rightarrow p_G(\mathbf{x})),$$

*where $p_G$ is a new predicate constant and $F'$ is the formula obtained from $F$ by replacing the occurrence of $QyG(y, \mathbf{x})$ with $p_G(\mathbf{x})$.*

*(c) Otherwise, if $Q$ is $\forall$ and the occurrence of $QyG(y, \mathbf{x})$ in $F$ is negative, then set $F$ to be the formula obtained from $F$ by replacing the occurrence of $QyG(y, \mathbf{x})$ with $\neg\exists y \neg G(y, \mathbf{x})$.*

We assume that the new predicate constants introduced by the translation do not belong to the signature of the input formula $F$. It is clear that this process terminates, and yields a formula that is quantifier-free. Since the number of times step (b) is applied is no more than the number of quantifiers in the input formula, and the new formulas added have the size polynomial to the input formula, it follows that the size of the resulting quantifier-free formula is polynomial in the size of the input formula.

The following theorem tells us that any almost universal sentence $F$ can be turned into the form $\forall \mathbf{x} G$, where $G$ is a quantifier-free formula. For any (second-order) sentences $F$ and $G$ of some signature and any subset $\sigma$ of that signature, we say that $F$ *is $\sigma$-equivalent to $G$*, denoted by $F \Leftrightarrow_\sigma G$, if the class of models of $F$ restricted to $\sigma$ is identical to the class of models of $G$ restricted to $\sigma$.

14. The maximality is understood here in terms of subformula relation. That is, we select a strictly positive occurrence of a subformula of $F$ of the form $\exists y H(y, \mathbf{x})$ that is not contained in any other subformula of $F$ of the same form.

**Theorem 6** *Let $F$ be a sentence of a signature $\sigma$, let $F'$ be the universal closure of the formula obtained from $F$ by applying translation* ELIM-QUANTIFIERS*, and let $\mathbf{q}$ be the list of new predicate constants introduced by the translation. If $F$ is almost universal relative to $\mathbf{p}$, then $\mathrm{SM}[F;\mathbf{p}]$ is $\sigma$-equivalent to $\mathrm{SM}[F';\mathbf{p},\mathbf{q}]$.*

The statement of the theorem becomes incorrect if we do not require $F$ to be almost universal relative to $\mathbf{p}$. For instance, if ELIM-QUANTIFIERS is applied to $\exists x\, p(x)$, it results in $\neg\neg q \wedge (p(x) \rightarrow q)$. However, $\mathrm{SM}[\exists x\, p(x);\ p]$ is not $\{p\}$-equivalent to $\mathrm{SM}[\forall x(\neg\neg q \wedge (p(x) \rightarrow q));\ p, q]$. The former is equivalent to saying that $p$ is a singleton. The latter is equivalent to $q \wedge \forall x \neg p(x) \wedge (q \leftrightarrow \exists x p(x))$, which is inconsistent.

### 5.3 F2LP: Computing Answer Sets of First-Order Formulas

Using translation ELIM-QUANTIFIERS defined in the previous section, we introduce translation F2LP that turns an almost universal formula into a logic program. We assume that the underlying signature contains finitely many predicate constants.

**Definition 8 (Translation F2LP)**
1. *Given a formula $F$ and a list of intensional predicates $\mathbf{p}$, apply translation* ELIM-QUANTIFIERS *(Definition 7) to $F$;*
2. *Add choice formulas $(q(\mathbf{x}) \vee \neg q(\mathbf{x}))$ for all non-intensional predicates $q$.*
3. *Turn the resulting quantifier-free formula into a logic program by applying the translation from Section 3 of the paper by Cabalar et al. (2005), which was also reviewed in Section 5.1.*

As explained in Section 5.1, due to the third step, this transformation may involve an exponential blowup in size. One can obtain a polynomial translation by replacing Step 3 with an alternative translation given in Section 4 of the paper by Cabalar et al.

The following theorem asserts the correctness of the translation.

**Theorem 7** *Let $F$ be a sentence of a signature $\sigma$, let $\mathbf{p}$ be a list of intensional predicates, and let $F'$ be the FOL representation of the program obtained from $F$ by applying translation* F2LP *with $\mathbf{p}$ as intensional predicates. If $F$ is almost universal relative to $\mathbf{p}$, then $\mathrm{SM}[F;\mathbf{p}]$ is $\sigma$-equivalent to*

$$\mathrm{SM}[F' \wedge \mathit{False}(\mathbf{p} \setminus pr(F'))].$$

**Example 4** *Consider one of the domain independent axioms in the discrete event calculus (DEC5 axiom):*

$$\begin{array}{l} \mathit{HoldsAt}(f,t) \wedge \neg \mathit{ReleasedAt}(f,t{+}1) \wedge \\ \qquad \neg \exists e(\mathit{Happens}(e,t) \wedge \mathit{Terminates}(e,f,t)) \rightarrow \mathit{HoldsAt}(f,t{+}1). \end{array} \tag{34}$$

*Step 1 of translation* F2LP *introduces the formula*

$$\mathit{Happens}(e,t) \wedge \mathit{Terminates}(e,f,t) \rightarrow q(f,t),$$

*and replaces (34) with*

$$\mathit{HoldsAt}(f,t) \wedge \neg \mathit{ReleasedAt}(f,t{+}1) \wedge \neg q(f,t) \rightarrow \mathit{HoldsAt}(f,t{+}1).$$

*Step 3 turns these formulas into rules*

$$q(f,t) \leftarrow Happens(e,t),\ Terminates(e,f,t)$$
$$HoldsAt(f,t{+}1) \leftarrow HoldsAt(f,t), not\ ReleasedAt(f,t{+}1),\ not\ q(f,t).$$

Turning the program obtained by applying translation F2LP into the input languages of LPARSE [15] and GRINGO [16] requires minor rewriting, such as moving equality and negated atoms in the head to the body [17] and adding domain predicates in the body for all variables occurring in the rule in order to reduce the many-sorted signature into the non-sorted one.[18]

System F2LP is an implementation of translation F2LP, which turns a first-order formula into the languages of LPARSE and GRINGO. The system can be downloaded from its home page

`http://reasoning.eas.asu.edu/f2lp` .

First-order formulas can be encoded in F2LP using the extended rule form $F \leftarrow G$, where $F$ and $G$ are first-order formulas that do not contain $\rightarrow$. The ASCII representation of the quantifiers and connectives are shown in the following table.

| Symbol | $\neg$ | $\sim$ | $\wedge$ | $\vee$ | $\leftarrow$ | $\bot$ | $\top$ | $\forall xyz$ | $\exists xyz$ |
|---|---|---|---|---|---|---|---|---|---|
| ASCII | `not` | `-` | `&` | `\|` | `<-` | `false` | `true` | `![X,Y,Z]:` | `?[X,Y,Z]:` |

For example, formula (34) can be encoded in the input language of F2LP as

```
holdsAt(F,T+1) <- holdsAt(F,T) & not releasedAt(F,T+1) &
                  not ?[E]:(happens(E,T) & terminates(E,F,T)).
```

The usual LPARSE and GRINGO rules (which have the rule arrow '`:-`') are also allowed in F2LP. Such rules are simply copied to the output. The program returned by F2LP can be passed to ASP grounders and solvers that accept LPARSE and GRINGO languages.

## 6. Computing the Event Calculus Using ASP Solvers

Using translation F2LP, we further turn the event calculus reformulation in Section 3.2 into answer set programs. The following procedure describes the process.

**Definition 9 (Translation EC2ASP)** *1. Given an event calculus description (16), rewrite all the definitional axioms of the form*

$$\forall \mathbf{x}(p(\mathbf{x}) \overset{def}{\leftrightarrow} G) \tag{35}$$

*in $\Xi$ as $\forall \mathbf{x}(G^{\neg\neg} \rightarrow p(\mathbf{x}))$, where $G^{\neg\neg}$ is obtained from $G$ by prepending $\neg\neg$ to all occurrences of intensional predicates Initiates, Terminates, Releases, Happens, $Ab_1, \dots, Ab_n$. Also prepend $\neg\neg$ to the strictly positive occurrences of the intensional predicates in the remaining axioms of $\Xi$. Let $\Xi'$ be the resulting formula obtained from $\Xi$.*

15. `http://www.tcs.hut.fi/Software/smodels`
16. `http://potassco.sourceforge.net`
17. For instance, `(X=Y) | -q(X,Y) :- p(X,Y)` is turned into `:- X!=Y, {not q(X,Y)}0, p(X,Y)`.
18. Alternatively this can be done by declaring variables using the `#domain` directive in LPARSE and GRINGO languages.

2. *Apply translation* F2LP *on* $\Sigma \wedge \Delta \wedge \Theta \wedge \Xi'$ *with the intensional predicates*

$$\{Initiates, Terminates, Releases, Happens, Ab_1, \ldots, Ab_n\} \cup \mathbf{p},$$

*where* $\mathbf{p}$ *is the set of all predicate constants* $p$ *in (35) as considered in Step 1.*

The following theorem states the correctness of the translation.

**Theorem 8** *Let* $T$ *be an event calculus description (16) of signature* $\sigma$ *that contains finitely many predicate constants, let* $F$ *be the FOL representation of the program obtained from* $T$ *by applying translation* EC2ASP*. Then* $T$ *is* $\sigma$*-equivalent to* $\text{SM}[F]$.

In view of the theorem, system F2LP can be used to compute event calculus descriptions by a simple rewriting as stated in translation EC2ASP.[19] The system can be used in place of the DEC reasoner in many existing applications of the event calculus, such as in robotics, security, video games, and web service composition, as listed in

`http://decreasoner.sourceforge.net/csr/decapps.html` .

The computational mechanism of the DEC reasoner is similar to our method as it is based on the reduction of event calculus reasoning to propositional satisfiability and uses efficient SAT solvers for computation. However, our method has some advantages.

First, it is significantly faster due to the efficient grounding mechanisms implemented in ASP systems. This is evidenced in some experiments reported in Appendix B.

Second, F2LP allows us to compute the full version of the event calculus, assuming that the domain is given and finite. On the other hand, the reduction implemented in the DEC reasoner is based on completion, which is weaker than circumscription. This makes the system unable to handle recursive axioms and disjunctive axioms, such as effect constraints and disjunctive event axioms (Section 3.1). For example, the DEC reasoner does not allow the following effect constraints which describe the indirect effects of an agent's walking on the objects that he is holding:

$$\begin{array}{l} HoldsAt(Holding(a,o),t) \wedge Initiates(e, InRoom(a,r),t) \\ \qquad\qquad\qquad\qquad \rightarrow Initiates(e, InRoom(o,r),t) \\ HoldsAt(Holding(a,o),t) \wedge Terminates(e, InRoom(a,r),t) \\ \qquad\qquad\qquad\qquad \rightarrow Terminates(e, InRoom(o,r),t). \end{array} \tag{36}$$

Third, we can enhance the event calculus reasoning by combining ASP rules with the event calculus description. In other words, the event calculus can be viewed as a high level action formalism on top of ASP. We illustrate this using the example from the work of Doğandağ, Ferraris, and Lifschitz (2004). There are 9 rooms and 12 doors as shown in Figure 1. Initially the robot "Robby" is in the middle room and all the doors are closed. The goal of the robot is to make all rooms accessible from each other. Figure 2 (File `robby`) shows an encoding of the problem in the language of F2LP. Atom `door(x,y)` denotes that there is a door between rooms `x` and `y`; `open(x,y)` denotes the event "Robby opening the door

19. Kim, Lee, and Palla (2009) presented a prototype of F2LP called ECASP that is tailored to the event calculus computation.

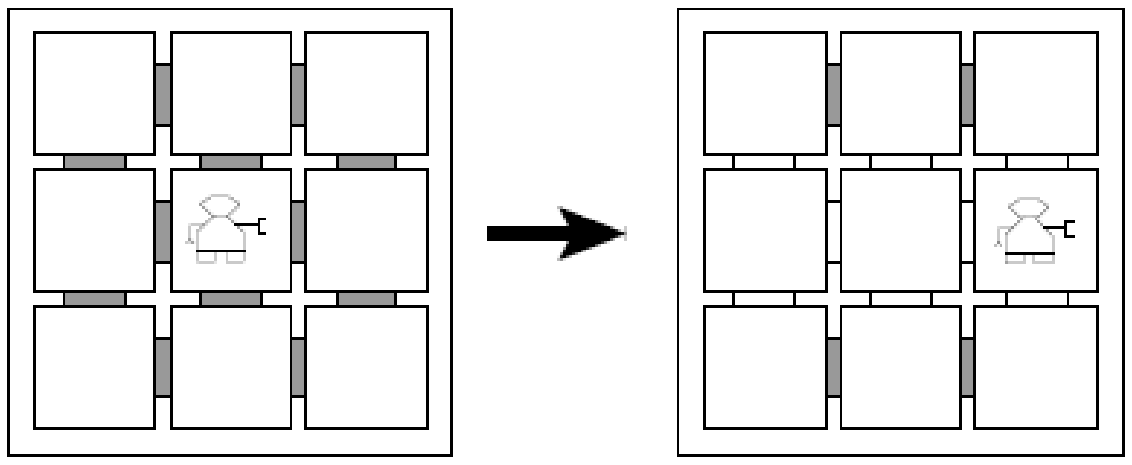

Figure 1: Robby's apartment in a 3 × 3 grid

between rooms `x` and `y`"; `goto(x)` denotes the event "Robby going to room `x`"; `opened(x, y)` denotes that the door between `x` and `y` has been opened; `inRoom(x)` denotes that Robby is in room `x`; `accessible(x, y)` denotes that `y` is accessible from `x`. Note that the rules defining the relation `accessible` are not part of event calculus axioms (Section 3.1). This example illustrates an advantage of allowing ASP rules in event calculus descriptions.

The minimal number of steps to solve the given problem is 11. We can find such a plan using the combination of F2LP, GRINGO (grounder) and CLASPD (solver for disjunctive programs) in the following way. [20]

```
$ f2lp dec robby | gringo -c maxstep=11 | claspD
```

File `dec` is an F2LP encoding of the domain independent axioms in the Discrete Event Calculus (The file is listed in Appendix A).[21] The following is one of the plans found:

```
happens(open(5,8),0) happens(open(5,2),1) happens(open(5,4),2)
happens(goto(4),3) happens(open(4,1),4) happens(open(4,7),5)
happens(goto(5),6) happens(open(5,6),7) happens(goto(6),8)
happens(open(6,9),9) happens(open(6,3),10)
```

## 7. Computing the Situation Calculus Using ASP Solvers

Using translation F2LP, we further turn the situation calculus reformulations in Sections 4.2 and 4.4 into answer set programs.

### 7.1 Representing Causal Action Theories by Answer Set Programs

The following theorem shows how to turn causal action theories into answer set programs.

**Theorem 9** *Let $\mathcal{D}$ be a finite causal action theory (23) of signature $\sigma$ that contains finitely many predicate constants, and let $F$ be the FOL representation of the program obtained by applying translation* F2LP *on*

$$\mathcal{D}_{caused} \wedge \mathcal{D}_{poss\rightarrow} \wedge \mathcal{D}^{-}_{rest} \wedge \mathcal{D}_{sit} \qquad (37)$$

*with the intensional predicates $\{Caused, Poss, Sit\}$. Then $\mathcal{D}$ is $\sigma$-equivalent to* $\mathrm{SM}[F]$.

20. One can use CLINGO instead of GRINGO and CLASPD if the output of F2LP is a nondisjunctive program.
21. The file is also available at `http://reasoning.eas.asu.edu/f2lp`, along with F2LP encodings of the domain independent axioms in other versions of the event calculus.

```
% File 'robby'

% objects
step(0..maxstep).
astep(0..maxstep-1) :- maxstep > 0.
room(1..9).

% variables
#domain step(T).
#domain room(R).
#domain room(R1).
#domain room(R2).

% position of the doors
door(R1,R2) <- R1 >= 1 & R2 >=1 & R1 < 4 & R2 < 4 & R2 = R1+1.
door(R1,R2) <- R1 >= 4 & R2 >= 4 & R1 < 7 & R2 < 7 & R2 = R1+1.
door(R1,R2) <- R1 >= 7 & R2 >= 7 & R1 < 10 & R2 < 10 & R2 = R1+1.
door(R1,R2) <- R2 < 10 & R2 = R1+3.

door(R1,R2) <- door(R2,R1).

% fluents
fluent(opened(R,R1)) <- door(R1,R2).
fluent(inRoom(R)).
% F ranges over the fluents
#domain fluent(F).

% events
event(open(R,R1)) <- door(R,R1).
event(goto(R)).
% E and E1 range over the events
#domain event(E).
#domain event(E1).

% effect axioms
initiates(open(R,R1),opened(R,R1),T).
initiates(open(R,R1),opened(R1,R),T).

initiates(goto(R2),inRoom(R2),T)
  <- holdsAt(opened(R1,R2),T) & holdsAt(inRoom(R1),T).

terminates(E,inRoom(R1),T)
  <- holdsAt(inRoom(R1),T) & initiates(E,inRoom(R2),T).

% action precondition axioms
holdsAt(inRoom(R1),T) <- happens(open(R1,R2),T).
```

```
% event occurrence constraint
not happens(E1,T) <- happens(E,T) & E != E1.

% state constraint
not holdsAt(inRoom(R2),T) <- holdsAt(inRoom(R1),T) & R1 != R2.

% accessibility
accessible(R,R1,T) <- holdsAt(opened(R,R1),T).
accessible(R,R2,T) <- accessible(R,R1,T) & accessible(R1,R2,T).

% initial state
not holdsAt(opened(R1,R2),0).
holdsAt(inRoom(5),0).

% goal state
not not accessible(R,R1,maxstep).

% happens is exempt from minimization in order to find a plan.
{happens(E,T)} <- T < maxstep.

% all fluents are inertial
not releasedAt(F,0).
```

Figure 2: Robby in F2LP

Similar to the computation of the event calculus in Section 6, the Herbrand stable models of (37) can be computed using F2LP and answer set solvers. The input to F2LP can be simplified as we limit attention to Herbrand models. We can drop axioms (18)–(21) as they are ensured by Herbrand models. Also, in order to ensure finite grounding, instead of $\mathcal{D}_{sit}$, we include the following set of rules $\Pi_{situation}$ in the input to F2LP.

```
nesting(0,s0).
nesting(L+1,do(A,S)) <- nesting(L,S) & action(A) & L < maxdepth.
situation(S) <- nesting(L,S).
final(S) <- nesting(maxdepth,S).
```

$\Pi_{situation}$ is used to generate finitely many situation terms whose depth is up to `maxdepth`, the value that can be given as an option in invoking GRINGO. Using the splitting theorem (Section C.1), it is not difficult to check that if a program $\Pi$ containing these rules has no occurrence of predicate `nesting` in any other rules and has no occurrence of predicate `situation` in the head of any other rules, then every answer set of $\Pi$ contains atoms $\mathtt{situation(do(a_m, do(a_{m-1}, do(\ldots, do(a_1, s0)))))}$ for all possible sequences of actions $\mathtt{a_1, \ldots, a_m}$ for $\mathtt{m} = \mathtt{0}, \ldots, \mathtt{maxdepth}$. Though this program does not satisfy syntactic conditions, such as $\lambda$-restricted (Gebser, Schaub, & Thiele, 2007), $\omega$-restricted (Syrjänen, 2004), or finite domain programs (Calimeri, Cozza, Ianni, & Leone, 2008), that answer set solvers usually impose in order to ensure finite grounding, the rules can still be finitely grounded

```
% File: suitcase
value(t).   value(f).   lock(l1).   lock(l2).

#domain value(V).
#domain lock(X).

fluent(up(X)).
fluent(open).

#domain fluent(F).
action(flip(X)).
#domain action(A).

depth(0..maxdepth).
#domain depth(L).

% defining the situation domain
nesting(0,s0).
nesting(L+1,do(A,S)) <- nesting(L,S) & L < maxdepth.
situation(S) <- nesting(L,S).
final(S) <- nesting(maxdepth,S).

% basic axioms
h(F,S) <- situation(S) & caused(F,t,S).
not h(F,S) <- situation(S) & caused(F,f,S).

% D_caused
caused(up(X),f,do(flip(X),S)) <-
   situation(S) & not final(S) & poss(flip(X),S) & h(up(X),S).

caused(up(X),t,do(flip(X),S)) <-
   situation(S) & not final(S) & poss(flip(X),S) & not h(up(X),S).

caused(open,t,S) <- situation(S) & h(up(l1),S) & h(up(l2),S).

% D_poss
poss(flip(X),S) <- situation(S).

% frame axioms
h(F,do(A,S)) <-
   h(F,S) & situation(S) & not final(S) & poss(A,S)
   & not ?[V]:caused(F,V,do(A,S)).

not h(F,do(A,S)) <-
   not h(F,S) & situation(S) & not final(S) & poss(A,S)
   & not ?[V]:caused(F,V,do(A,S)).

% h is non-intensional.
{h(F,S)} <- situation(S).
```

Figure 3: Lin's Suitcase in the language of f2lp

by gringo Version 3.$x$, which does not check such syntactic conditions.[22] It is not difficult to see why the program above leads to finite grounding since we provide an explicit upper limit for the nesting depth of function `do`.

In addition to $\Pi_{situation}$, we use the following program $\Pi_{executable}$ in order to represent the set of executable situations (Reiter, 2001):

```
executable(s0).
executable(do(A,S)) <- executable(S) & poss(A,S) & not final(S)
                      & situation(S) & action(A).
```

Figure 3 shows an encoding of Lin's suitcase example (1995) in the language of f2lp (`h` is used to represent *Holds*), which describes a suitcase that has two locks and a spring loaded mechanism which will open the suitcase when both locks are up. This example illustrates how the ramification problem is handled in causal action theories. Since we fix the domain of situations to be finite, we require that actions not be effective in the final situations. This is done by introducing atom `final(S)`.

Consider the simple temporal projection problem by Lin (1995). Initially the first lock is down and the second lock is up. What will happen if the first lock is flipped? Intuitively, we expect both locks to be up and the suitcase to be open. We can automate the reasoning by using the combination of f2lp, gringo and claspD. First, we add $\Pi_{executable}$ and the following rules to the theory in Figure 3. In order to check if the theory entails that flipping the first lock is executable, and that the suitcase is open after the action, we encode the negation of these facts in the last rule.

```
% initial situation
<- h(up(l1),s0).
h(up(l2),s0).

% query
<- executable(do(flip(l1),s0)) & h(open,do(flip(l1),s0)).
```

We check the answer to the temporal projection problem by running the command:

```
$ f2lp suitcase | gringo -c maxdepth=1 | claspD
```

claspD returns no answer set as expected.

Now, consider a simple planning problem for opening the suitcase when both locks are initially down. We add $\Pi_{executable}$ and the following rules to the theory in Figure 3. The last rule encodes the goal.

```
% initial situation
<- h(up(l1),s0).
<- h(up(l2),s0).
<- h(open,s0).

% goal
<- not ?[S]: (executable(S) & h(open,S)).
```

When `maxdepth` is 1, the combined use of f2lp, gringo and claspD results in no answer sets, and when `maxdepth` is 2, it finds the unique answer set that contains both

22. Similarly, system dlv-complex allows us to turn off the finite domain checking (option `-nofdcheck`). That system was used in a conference paper (Lee & Palla, 2010) that this article is based on.

h(open, do(flip(l2), do(flip(l1), s0))) and h(open, do(flip(l1), do(flip(l2), s0))), each of which encodes a plan. In other words, the single answer set encodes multiple plans in different branches of the situation tree, which allows us to combine information about the different branches in one model. This is an instance of hypothetical reasoning that is elegantly handled in the situation calculus due to its branching time structure. Belleghem, Denecker, and Schreye (1997) note that the linear time structure of the event calculus is limited to handle such hypothetical reasoning allowed in the situation calculus.

### 7.2 Representing Basic Action Theories by Answer Set Programs

Since a BAT $T$ (not including the second-order axiom (22)) can be viewed as a first-order theory under the stable model semantics with the list of intensional predicates being empty, it follows that F2LP can be used to turn $T$ into a logic program. As before, we focus on ASP-style BAT.

**Theorem 10** *Let $T$ be a ASP-style BAT (26) of signature $\sigma$ that contains finitely many predicate constants, and let $F$ be the FOL representation of the program obtained by applying translation* F2LP *on $T$ with intensional predicates $\{Holds, \sim Holds, Poss\}$. Then* $\text{SM}[T; Holds, {\sim}Holds, Poss]$ *is $\sigma$-equivalent to* $\text{SM}[F;\ \sigma(F) \cup \{Poss\}]$.

Figure 4 shows an encoding of the "broken object" example discussed by Reiter (1991). Consider the simple projection problem of determining if an object $o$, which is next to bomb $b$, is broken after the bomb explodes. We add $\Pi_{executable}$ and the following rules to the theory in Figure 4.

```
% initial situation
not h(broken(o),s0) & h(fragile(o),s0) & h(nexto(b,o),s0).
not h(holding(p,o),s0) & not h(exploded(b),s0).

% query
<- executable(do(explode(b),s0)) & h(broken(o),do(explode(b),s0)).
```

The command

```
$ f2lp broken | gringo -c maxdepth=1 | claspD
```

returns no answer set as expected.

## 8. Related Work

Identifying a syntactic class of theories on which different semantics coincide is important in understanding the relationship between them. It is known that, for tight logic programs and tight first-order formulas, the stable model semantics coincides with the completion semantics (Fages, 1994; Erdem & Lifschitz, 2003; Ferraris et al., 2011). This fact helps us understand the relationship between the two semantics, and led to the design of the answer set solver CMODELS-1 [23] that computes answer sets using completion. Likewise the class of canonical formulas introduced here helps us understand the relationship between the stable model semantics and circumscription. The class of canonical formulas is the largest

23. http://www.cs.utexas.edu/users/tag/cmodels

```
% File: broken
% domains other than situations
person(p).    object(o).    bomb(b).

#domain person(R).
#domain object(Y).
#domain bomb(B).

fluent(holding(R,Y)).    fluent(nexto(B,Y)).     fluent(fragile(Y)).
fluent(broken(Y)).       fluent(exploded(B)).

action(drop(R,Y)).       action(explode(B)).     action(repair(R,Y)).

#domain fluent(F).
#domain action(A).

depth(0..maxdepth).
#domain depth(L).

% defining the situation domain
nesting(0,s0).
nesting(L+1,do(A,S)) <- nesting(L,S) & L < maxdepth.
situation(S) <- nesting(L,S).
final(S) <- nesting(maxdepth,S).

% Effect Axioms
h(broken(Y),do(drop(R,Y),S)) <- situation(S) & h(fragile(Y),S) & not final(S).
h(broken(Y),do(explode(B),S)) <- situation(S) & h(nexto(B,Y),S) & not final(S).
h(exploded(B),do(explode(B),S)) <- situation(S) & not final(S).
-h(broken(Y),do(repair(R,Y),S)) <- situation(S) & not final(S).
-h(holding(R,Y),do(drop(R,Y),S)) <- situation(S) & not final(S).

% Action precondition axioms
poss(drop(R,Y),S) <- h(holding(R,Y),S) & situation(S).
poss(explode(B),S) <- situation(S) & not h(exploded(B),S).
poss(repair(R,Y),S) <- situation(S) & h(broken(Y),S).

% inertial axioms
h(F,do(A,S)) <- h(F,S) & not -h(F,do(A,S)) & situation(S) & not final(S).
-h(F,do(A,S)) <- -h(F,S) & not h(F,do(A,S)) & situation(S) & not final(S).

% D_exogeneous_0
h(F,s0) | -h(F,s0).

% Consider only those interpretations that are complete on Holds
<- not h(F,S) & not -h(F,S) & situation(S).
```

Figure 4: Broken object example in the language of F2LP

syntactic class of first-order formulas identified so far on which the stable models coincide with the models of circumscription. In other words, minimal model reasoning and stable model reasoning are indistinguishable on canonical formulas.

Proposition 8 from the work of Lee and Lin (2006) shows an embedding of propositional circumscription in logic programs under the stable model semantics. Our theorem on canonical formulas is a generalization of this result to the first-order case. Janhunen and Oikarinen (2004) showed another embedding of propositional circumscription in logic programs, and implemented the system CIRC2DLP,[24] but their translation appears quite different from the one by Lee and Lin.

Zhang, Zhang, Ying, and Zhou (2011) show an embedding of first-order circumscription in first-order stable model semantics. Theorem 3 from that paper is reproduced as follows.[25]

**Theorem 11** *(Zhang et al., 2011, Thm. 3) Let $F$ be a formula in negation normal form and let $\mathbf{p}$ be a finite list of predicate constants. Let $F^{\neg\neg}$ be the formula obtained from $F$ by replacing every $p(\mathbf{t})$ by $\neg\neg p(\mathbf{t})$, and let $F^c$ be the formula obtained from $F$ by replacing every $\neg p(\mathbf{t})$ by $p(\mathbf{t}) \rightarrow Choice(\mathbf{p})$, where $p$ is in $\mathbf{p}$ and $\mathbf{t}$ is a list of terms. Then* $\mathrm{CIRC}[F; \mathbf{p}]$ *is equivalent to* $\mathrm{SM}[F^{\neg\neg} \wedge F^c; \mathbf{p}]$.

In comparison with Theorem 1, this theorem can be applied to characterize circumscription of arbitrary formulas in terms of stable models by first rewriting the formulas into negation normal form. While Theorem 1 is applicable to canonical formulas only, it does not require any transformation, and the characterization is bidirectional in the sense that it can be also viewed as a characterization of stable models in terms of circumscription.

Zhang et al. (2011) also introduce a translation that turns arbitrary first-order formulas into logic programs, but this work is limited to finite structures only. On the other hand, our translation F2LP (Definition 8) works for almost universal formulas only, but is not limited to finite structures.

The situation calculus and the event calculus are widely studied action formalisms, and there are several papers that compare and relate them (e.g., Belleghem, Denecker, & Schreye, 1995; Provetti, 1996; Belleghem et al., 1997; Kowalski & Sadri, 1997).

Prolog provides a natural implementation for basic action theories since definitional axioms can be represented by Prolog rules according to the Clark's theorem (Reiter, 2001, Chapter 5). The Lloyd-Topor transformation that is used to turn formulas into Prolog rules is similar to translation F2LP, but the difference is that the former preserves the completion semantics and the latter preserves the stable model semantics.

Lin and Wang (1999) describe a language that can be used to represent a syntactically restricted form of Lin's causal situation calculus, called "clausal causal theories," which does not allow quantifiers. They show how to translate that language into answer set programs with strong negation, the answer sets of which are then used to obtain fully instantiated successor state axioms and action precondition axioms. This is quite different from our approach, which computes the propositional models of the full situation calculus theories directly.

Kautz and Selman (1992) introduce linear encodings that are similar to a propositionalized version of the situation calculus (McCarthy & Hayes, 1969). Lin (2003) introduces

24. `http://www.tcs.hut.fi/Software/circ2dlp`
25. This is a bit simpler than the original statement because some redundancy is dropped.

an action description language and describes a procedure to compile an action domain in that language into a complete set of successor state axioms, from which a STRIPS-like description can be extracted. The soundness of the procedure is shown with respect to a translation from action domain descriptions into Lin's causal action theories. However, that procedure is based on completion and as such cannot handle recursive axioms unlike our approach.

Denecker and Ternovska (2007) present an inductive variant of the situation calculus represented in ID-logic (Denecker & Ternovska, 2008)—classical logic extended with inductive definitions. ID-logic and the first-order stable model semantics appear to be closely related, but the precise relationship between them has yet to be shown.

## 9. Conclusion

The first-order stable model semantics is defined similar to circumscription. This paper takes advantage of that definition to identify a class of formulas on which minimal model reasoning and stable model reasoning coincide, and uses this idea to reformulate the situation calculus and the event calculus in the first-order stable model semantics. Together with the translation that turns an almost universal sentence into a logic program, we show that reasoning in the situation calculus and the event calculus can be reduced to computing answer sets. We implemented system F2LP, a front-end to ASP solvers that allows us to compute these circumscriptive action theories. The mathematical tool sets and the system presented in this paper may also be useful in relating other circumscriptive theories to logic programs. Also, the advances in ASP solvers may improve the computation of circumscriptive theories.

## Acknowledgments

We are grateful to Yuliya Lierler, Vladimir Lifschitz, Erik Mueller, Heng Zhang, Yan Zhang, and the anonymous referees for their useful comments and discussions. The authors were partially supported by the National Science Foundation under Grant IIS-0916116.

## Appendix A. File 'dec' in the Language of F2LP

File 'dec' encodes the domain independent axioms in the discrete event calculus. This file is to be used together with event calculus domain descriptions as shown in Section 6.

```
% File 'dec'

#domain fluent(F).
#domain fluent(F1).
#domain fluent(F2).
#domain event(E).
#domain time(T).
#domain time(T1).
#domain time(T2).

time(0..maxstep).
```

```
% DEC 1
stoppedIn(T1,F,T2) <- happens(E,T) & T1<T & T<T2 & terminates(E,F,T).

% DEC 2
startedIn(T1,F,T2) <- happens(E,T) & T1<T & T<T2 & initiates(E,F,T).

% DEC 3
holdsAt(F2,T1+T2) <- happens(E,T1) & initiates(E,F1,T1) & T2>0 &
   trajectory(F1,T1,F2,T2) & not stoppedIn(T1,F1,T1+T2) & T1+T2<=maxstep.

% DEC 4
holdsAt(F2,T1+T2) <- happens(E,T1) & terminates(E,F1,T1) & 0<T2 &
   antiTrajectory(F1,T1,F2,T2) & not startedIn(T1,F1,T1+T2) &
   T1+T2<=maxstep.

% DEC 5
holdsAt(F,T+1) <- holdsAt(F,T) & not releasedAt(F,T+1) &
   not ?[E]:(happens(E,T) & terminates(E,F,T)) & T<maxstep.

% DEC 6
not holdsAt(F,T+1) <- not holdsAt(F,T) & not releasedAt(F,T+1) &
   not ?[E]:(happens(E,T) & initiates(E,F,T)) & T<maxstep.

% DEC 7
releasedAt(F,T+1) <-
   releasedAt(F,T) & not ?[E]:(happens(E,T) &
   (initiates(E,F,T) | terminates(E,F,T))) & T<maxstep.

% DEC 8
not releasedAt(F,T+1) <- not releasedAt(F,T) &
   not ?[E]: (happens(E,T) & releases(E,F,T)) & T<maxstep.

% DEC 9
holdsAt(F,T+1) <- happens(E,T) & initiates(E,F,T) & T<maxstep.

% DEC 10
not holdsAt(F,T+1) <- happens(E,T) & terminates(E,F,T) & T<maxstep.

% DEC 11
releasedAt(F,T+1) <- happens(E,T) & releases(E,F,T) & T<maxstep.

% DEC 12
not releasedAt(F,T+1) <- happens(E,T) &
   (initiates(E,F,T) | terminates(E,F,T)) & T<maxstep.

{holdsAt(F,T)}.
{releasedAt(F,T)}.
```

| Problem (max. step) | DEC reasoner | DEC reasoner (minisat) | F2LP with LPARSE + CMODELS | F2LP with GRINGO + CMODELS | F2LP with GRINGO + CLASP(D) | F2LP with CLINGO |
|---|---|---|---|---|---|---|
| BusRide (15) | — | — | 0.04s (0.03s + 0.01s) A:902/R:7779 C:0 | 0.00s (0.00s + 0.00s) A:355/R:555 C:0 | 0.01s (0.00s + 0.01s) A:448/R:647 | — |
| Commuter (15) | — | — | 77.29s (45.74s + 31.55s) A:32861/R:8734019 C:0 | 0.15s (0.07s + 0.08s) A:5269/R:24687 C:5308 | 0.2s (0.07s + 0.13s) A:13174/R:24687 | 0.14s |
| Kitchen Sink (25) | 39.0s (38.9s + 0.1s) A:1014/C:12109 | 38.9s (38.9s + 0.00s) A:1014/C:12109 | 6.19s (2.99s + 3.20s) A:121621/R:480187 C:0 | 0.44s (0.19s + 0.25s) A:11970/R:61932 C:0 | 0.24s (0.18s + 0.06s) A:11970/R:61932 | 0.20s |
| Thielscher Circuit (40) | 6.5s (6.3s + 0.2s) A:1394/C:42454 | 6.3s (6.3s + 0.0s) A:1394/C:42454 | 0.42s (0.27s + 0.15s) A:9292/R:53719 C:0 | 0.19s (0.09s + 0.1s) A:4899/R:35545 C:0 | 0.12s (0.09s + 0.03s) A:4899/R:35545 | 0.1s |
| Walking Turkey (15) | — | — | 0.00s (0.00s + 0.00s) A:370/R:518 C:0 | 0.00s (0.00s + 0.00s) A:316/R:456 C:0 | 0.00s (0.00s + 0.00s) A:316/R:456 | 0.00s |
| Falling w/ AntiTraj (15) | 141.8s (141.4s + 0.4s) A:416/C:3056 | 141.7s (141.7s + 0.00s) A:416/C:3056 | 0.08s (0.05s + 0.03s) A:4994/R:9717 C:0 | 0.04s (0.02s + 0.02s) A:3702/R:7414 C:0 | 0.03s (0.03s + 0.00s) A:3702/R:7414 | 0.03s |
| Falling w/ Events (25) | 59.5s (59.5s + 0.0s) A:1092/C:12351 | 59.4s (59.4s + 0.0s) A:1092/C:12351 | 4.95s (2.57s + 2.38s) A:1240/R:388282 C:1436 | 0.46s (0.20s + 0.26s) A:1219/R:71266 C:1415 | 0.28s (0.20s + 0.08s) A:13829/R:71266 | 0.22s |
| HotAir Baloon (15) | 32.2s (32.2s + 0.0s) A:288/C:1163 | 32.3s (32.3s + 0.0s) A:288/C:1163 | 0.01s (0.01s + 0.00s) A:494/R:2451 C:689 | 0.0s (0.0s + 0.0s) A:492/R:1835 C:681 | 0.0s (0.0s + 0.0s) A:1063/R:1835 | 0.01s |
| Telephone1 (40) | 9.3s (9.2s + 0.1s) A:5419/C:41590 | 9.1s (9.1s + 0.0s) A:5419/C:41590 | 0.22s (0.13s + 0.09s) A:21414/R:27277 C:0 | 0.11s (0.08s + 0.03s) A:9455/R:13140 C:0 | 0.07s (0.06s + 0.01s) A:9455/R:13140 | 0.07s |

A: number of atoms, C: number of clauses, R: number of ground rules

Figure 5: Comparing the DEC reasoner and F2LP with answer set solvers

## Appendix B. Comparing the DEC Reasoner with ASP-based Event Calculus Reasoner

We compared the performance of the DEC reasoner (v 1.0) running RELSAT (v 2.2) and MINISAT (v 2.2) with the following:

- F2LP (v 1.11) with LPARSE (v 1.0.17)+CMODELS (v 3.79) running MINISAT (v 2.0 beta),
- F2LP (v 1.11) with GRINGO (v 3.0.3)+CMODELS (v 3.79) running MINISAT (v 2.0 beta),
- F2LP (v 1.11) with GRINGO (v 3.0.3) +CLASP (v 2.0.2) (CLASPD (v 1.1.2) used instead for disjunctive programs), and
- F2LP (v 1.11) with CLINGO (v 3.0.3 (CLASP v 1.3.5)).

F2LP turns an input theory into the languages of LPARSE and GRINGO, and LPARSE and GRINGO turn the result into a ground ASP program. CMODELS turns this ground program into a set of clauses and then invokes a SAT solver to compute answer sets, while CLASP computes answer sets using the techniques similar to those used in SAT solvers. CLINGO is a system that combines GRINGO and CLASP in a monolithic way.

The first five examples in Figure 5 are part of the benchmark problems from the work of Shanahan (1997, 1999). The next four are by Mueller (2006). (We increased timepoints

| Problem (max. step) | DEC reasoner (MINISAT) | F2LP with GRINGO + CMODELS | F2LP with GRINGO + CLASP |
|---|---|---|---|
| ZooTest1 (16) | > 2h | 50.48s (6.66s + 43.82s) A:930483/R:2272288 C:3615955 | 29.01s (6.66s + 22.35s) A:153432/R:2271175 |
| ZooTest2 (22) | > 2h | 159.51s (12.36s + 147.15s) A:2241512/R:4153670 C:8864228 | 210.55s (12.36s + 198.19s) A:219220/R:4152137 |
| ZooTest3 (23) | > 2h | 142.68s (13.55s + 129.13s) A:2505940/R:4556928 C:9914568 | 196.63s (13.55s + 183.08s) A:230731/R:4555325 |

A: number of atoms, C: number of clauses, R: number of ground rules to see more notable differences.) More examples can be found from the F2LP homepage. All experiments were done on a Pentium machine with 3.06 GHz CPU and 4GB RAM running 64 bit Linux. The reported run times are in seconds and were obtained using the Linux `time` command ("user time + sys time"), except for the DEC reasoner for which we recorded the times reported by the system. This was for fair comparisons in order to avoid including the time spent by the DEC reasoner in producing output in a neat format, which sometimes takes non-negligible time. For the DEC reasoner, the times in parentheses are "(SAT encoding time + SAT solving time)." For the others, they are the times spent by each of the grounder and the solver. CMODELS time includes the time spent in converting the ground program generated by LPARSE/GRINGO into a set of clauses, and calling the SAT solver. The time spent by F2LP in translating an event calculus description into an answer set program (retaining variables) is negligible for these problems. '—' denotes that the system cannot solve the example due to the limited expressivity. For instance, `BusRide` includes disjunctive event axioms, which results in a disjunctive program that cannot be handled by CLINGO. Similarly, the DEC reasoner cannot handle `BusRide` (disjunctive event axioms), `Commuter` (compound events) and `Walking Turkey` (effect constraints). As is evident from the experiments, the main reason for the efficiency of the ASP-based approach is the efficient grounding mechanisms implemented in the ASP grounders. Though the DEC reasoner and CMODELS call the same SAT solver MINISAT, the number of atoms processed by the DEC reasoner is in general much smaller. This is because the DEC reasoner adopts an optimized encoding method (that is based on predicate completion) which avoids a large number of ground instances of atoms such as $\mathit{Initiates}(e, f, t)$, $\mathit{Terminates}(e, f, t)$, and $\mathit{Releases}(e, f, t)$ (Mueller, 2004, Section 4.4). On the other hand, in several examples, the number of clauses generated by CMODELS is 0, which means that the answer sets were found without calling the SAT solver. This is because for these examples the unique answer set coincides with the well-founded model, which is efficiently computed by CMODELS in a preprocessing step before calling SAT solvers. Out of the 14 benchmark examples by Shanahan (1997, 1999), 10 of them belong to this case when LPARSE is used for grounding.

Figure 6: Zoo World in DEC reasoner and ASP

In the experiments in Figure 5, the solving times are negligible for most of the problems. We also experimented with some computationally hard problems, where solving takes more time than grounding. Figure 6 shows runs of a medium-size action domain, the Zoo World (Akman, Erdoğan, Lee, Lifschitz, & Turner, 2004). All the tests shown in the table are planning problems where *max. step* is the length of a minimal plan. The cut-off time was 2 hours and the DEC reasoner did not terminate within that time for any of the problems. In fact, the entire time was spent on SAT encoding and the SAT solver was never called. On the other hand, the ASP grounder GRINGO took only a few seconds to ground the domain and, unlike in Figure 5, the solvers took much more time than the grounder. As we can see, CMODELS with MINISAT performed better than CLASP on two of the problems. To check the time taken by MINISAT on the encoding generated by the DEC reasoner, we ran ZooTest1 to completion. The DEC reasoner terminated after 116578.1 seconds (32.38 hours).

## Appendix C. Proofs

### C.1 Review of Some Useful Theorems

We review some theorems by Ferraris et al. (2011) and Ferraris et al. (2009) which will be used to prove our main results. In fact, we will provide a version of the splitting theorem which is slightly more general than the one given by Ferraris et al. (2009), in order to facilitate our proof efforts.

**Lemma 1** *Formula*

$$\mathbf{u} \leq \mathbf{p} \rightarrow ((\neg F)^*(\mathbf{u}) \leftrightarrow \neg F)$$

*is logically valid.*

**Theorem 12** *(Ferraris et al., 2011, Thm. 2) For any first-order formula $F$ and any disjoint lists $\mathbf{p}$, $\mathbf{q}$ of distinct predicate constants,*

$$\text{SM}[F; \mathbf{p}] \leftrightarrow \text{SM}[F \wedge \mathit{Choice}(\mathbf{q});\ \mathbf{p} \cup \mathbf{q}]$$

*is logically valid.*

Let $F$ be a first-order formula. A *rule* of $F$ is an implication that occurs strictly positively in $F$. The *predicate dependency graph* of $F$ (relative to $\mathbf{p}$) is the directed graph that

- has all members of $\mathbf{p}$ as its vertices, and
- has an edge from $p$ to $q$ if, for some rule $G \rightarrow H$ of $F$,
  - $p$ has a strictly positive occurrence in $H$, and
  - $q$ has a positive occurrence in $G$ that does not belong to any subformula of $G$ that is negative on $\mathbf{p}$.

**Theorem 13** *(Ferraris et al., 2009, Splitting Thm.) Let $F$, $G$ be first-order sentences, and let $\mathbf{p}$, $\mathbf{q}$ be finite disjoint lists of distinct predicate constants. If*

*(a) each strongly connected component of the predicate dependency graph of* $F \wedge G$ *relative to* $\mathbf{p}$, $\mathbf{q}$ *is either a subset of* $\mathbf{p}$ *or a subset of* $\mathbf{q}$*,*

*(b)* $F$ *is negative on* $\mathbf{q}$*, and*

*(c)* $G$ *is negative on* $\mathbf{p}$

*then*

$$\text{SM}[F \wedge G;\ \mathbf{p} \cup \mathbf{q}] \leftrightarrow \text{SM}[F;\ \mathbf{p}] \wedge \text{SM}[G;\ \mathbf{q}]$$

*is logically valid.*

The theorem is slightly more general than the one by Ferraris et al. (2009) in that the notion of a dependency graph above yields less edges than the one given by Ferraris et al. Instead of

– $q$ has a positive occurrence in $G$ that does not belong to any subformula of $G$ that is negative on $\mathbf{p}$,

Ferraris et al.'s definition has

– $q$ has a positive occurrence in $G$ that does not belong to any subformula of the form $\neg K$.

For instance, according to Ferraris et al., the dependency graph of

$$((p \rightarrow q) \rightarrow r) \rightarrow p \tag{38}$$

relative to $p$ has two edges (from $p$ to $r$, and from $p$ to $p$), while the dependency graph according to our definition has no edges.

On the other hand, the generalization is not essential in view of the following theorem.

**Theorem 14** *(Ferraris et al., 2009, Thm. on Double Negations) Let* $H$ *be a sentence,* $F$ *a subformula of* $H$*, and* $H^{\neg\neg}$ *the sentence obtained from* $H$ *by inserting* $\neg\neg$ *in front of* $F$*. If the occurrence of* $F$ *is* $\mathbf{p}$*-negated in* $H$*, then* $\text{SM}[H; \mathbf{p}]$ *is equivalent to* $\text{SM}[H^{\neg\neg}; \mathbf{p}]$*.*

For instance, $\text{SM}[(38); p]$ is equivalent to $\text{SM}[\neg\neg((p \rightarrow q) \rightarrow r) \rightarrow p;\ p]$. The dependency graph of $\neg\neg((p \rightarrow q) \rightarrow r) \rightarrow p$ relative to $p$ according to the definition by Ferraris et al. is identical to the dependency graph of (38) relative to $p$ according to our definition.

Next, we say that a formula $F$ is in *Clark normal form* (relative to the list $\mathbf{p}$ of intensional predicates) if it is a conjunction of sentences of the form

$$\forall \mathbf{x}(G \rightarrow p(\mathbf{x})), \tag{39}$$

one for each intensional predicate $p$, where $\mathbf{x}$ is a list of distinct object variables, and $G$ has no free variables other than those in $\mathbf{x}$. The *completion* (relative to $\mathbf{p}$) of a formula $F$ in Clark normal form is obtained by replacing each conjunctive term (39) with

$$\forall \mathbf{x}(p(\mathbf{x}) \leftrightarrow G).$$

The following theorem relates SM to completion. We say that $F$ is *tight* on $\mathbf{p}$ if the predicate dependency graph of $F$ relative to $\mathbf{p}$ is acyclic.

**Theorem 15** *(Ferraris et al., 2011) For any formula* $F$ *in Clark normal form that is tight on* $\mathbf{p}$*, formula* $\text{SM}[F; \mathbf{p}]$ *is equivalent to the completion of* $F$ *relative to* $\mathbf{p}$*.*

### C.2 Proof of Proposition 1

Using Theorem 12 and Theorem 13,

$$\begin{aligned}
\mathrm{SM}[F;\mathbf{p}] &\Leftrightarrow \mathrm{SM}[F;\ \mathbf{p}\cap pr(F)] \wedge \mathrm{SM}[\top;\ \mathbf{p}\backslash pr(F)]\\
&\Leftrightarrow \mathrm{SM}[F;\ \mathbf{p}\cap pr(F)] \wedge \mathit{False}(\mathbf{p}\backslash pr(F))\\
&\Leftrightarrow \mathrm{SM}[F \wedge \mathit{Choice}(pr(F)\backslash \mathbf{p})] \wedge \mathit{False}(\mathbf{p}\backslash pr(F))\\
&\Leftrightarrow \mathrm{SM}[F \wedge \mathit{Choice}(pr(F)\backslash \mathbf{p}) \wedge \mathit{False}(\mathbf{p}\backslash pr(F))].
\end{aligned}$$

□

### C.3 Proof of Theorem 1

In the following, $F$ is a formula, $\mathbf{p}$ is a list of distinct predicate constants $p_1, \dots, p_n$, and $\mathbf{u}$ is a list of distinct predicate variables $u_1, \dots, u_n$ of the same length as $\mathbf{p}$.

**Lemma 2** *(Ferraris et al., 2011, Lemma 5) Formula*

$$\mathbf{u} \leq \mathbf{p} \rightarrow (F^*(\mathbf{u}) \rightarrow F)$$

*is logically valid.*

**Lemma 3** *If every occurrence of every predicate constant from* $\mathbf{p}$ *is strictly positive in* $F$*,*

$$(\mathbf{u} \leq \mathbf{p}) \rightarrow (F^*(\mathbf{u}) \leftrightarrow F(\mathbf{u}))$$

*is logically valid.*

**Proof**. By induction. We will show only the case when $F$ is $G \rightarrow H$. The other cases are straightforward. Consider

$$F^*(\mathbf{u}) = (G^*(\mathbf{u}) \rightarrow H^*(\mathbf{u})) \wedge (G \rightarrow H).$$

Since every occurrence of predicate constants from $\mathbf{p}$ in $F$ is strictly positive, $G$ contains no predicate constants from $\mathbf{p}$, so that $G^*(\mathbf{u})$ is equivalent to $G(\mathbf{u})$, which is the same as $G$. Also, by I.H., $H^*(\mathbf{u}) \leftrightarrow H(\mathbf{u})$ is logically valid. Therefore it is sufficient to prove that under the assumption $\mathbf{u} \leq \mathbf{p}$,

$$(G \rightarrow H(\mathbf{u})) \wedge (G \rightarrow H) \leftrightarrow (G \rightarrow H(\mathbf{u}))$$

is logically valid. From left to right is clear. Assume $(\mathbf{u} \leq \mathbf{p})$, $G \rightarrow H(\mathbf{u})$, and $G$. We get $H(\mathbf{u})$, which is equivalent to $H^*(\mathbf{u})$ by I.H. By Lemma 2, we conclude $H$. □

The proof of Theorem 1 is immediate from the following lemma, which can be proved by induction.

**Lemma 4** *If* $F$ *is canonical relative to* $\mathbf{p}$*, then formula*

$$(\mathbf{u} \leq \mathbf{p}) \wedge F \rightarrow (F^*(\mathbf{u}) \leftrightarrow F(\mathbf{u}))$$

*is logically valid.*

**Proof**.

- $F$ is an atomic formula. Trivial.
- $F = G \wedge H$. Follows from I.H.
- $F = G \vee H$. Assume $(\mathbf{u} \leq \mathbf{p}) \wedge (G \vee H)$. Since $G \vee H$ is canonical relative to $\mathbf{p}$, every occurrence of every predicate constant from $\mathbf{p}$ is strictly positive in $G$ or in $H$, so that, by Lemma 3, $G^*(\mathbf{u})$ is equivalent to $G(\mathbf{u})$, and $H^*(\mathbf{u})$ is equivalent to $H(\mathbf{u})$.
- $F = G \rightarrow H$. Assume $(\mathbf{u} \leq \mathbf{p}) \wedge (G \rightarrow H)$. It is sufficient to show

$$(G^*(\mathbf{u}) \rightarrow H^*(\mathbf{u})) \leftrightarrow (G(\mathbf{u}) \rightarrow H(\mathbf{u})). \tag{40}$$

  Since $G \rightarrow H$ is canonical relative to $\mathbf{p}$, every occurrence of every predicate constant from $\mathbf{p}$ in $G$ is strictly positive in $G$, so that, by Lemma 3, $G^*(\mathbf{u})$ is equivalent to $G(\mathbf{u})$.

  - *Case 1:* $\neg G$. By Lemma 2, $\neg G^*(\mathbf{u})$. The claim follows since $\neg G^*(\mathbf{u})$ is equivalent to $\neg G(\mathbf{u})$.
  - *Case 2:* $H$. By I.H. $H^*(\mathbf{u})$ is equivalent to $H(\mathbf{u})$. The claim follows since $G^*(\mathbf{u})$ is equivalent to $G(\mathbf{u})$.
- $F = \forall x G$. Follows from I.H.
- $F = \exists x G$. Since every occurrence of every predicate constant from $\mathbf{p}$ in $G$ is strictly positive in $G$, the claim follows from Lemma 3.

□

### C.4 Proof of Theorem 2

**Proof**. *Between (a) and (b):* Follows immediately from Theorem 1.

*Between (b) and (c):* Note first that $\Xi$ is equivalent to $\text{SM}[\Xi; \emptyset]$. Since

- every strongly connected component in the dependency graph of $\Sigma \wedge \Delta$ relative to $\{I, T, R, H\}$ either belongs to $\{I, T, R\}$ or $\{H\}$,
- $\Sigma$ is negative on $\{H\}$, and
- $\Delta$ is negative on $\{I, T, R\}$,

it follows from Theorem 13 that (b) is equivalent to

$$\text{SM}[\Sigma \wedge \Delta;\ I, T, R, H] \wedge \text{SM}[\Theta;\ Ab_1, \dots, Ab_n] \wedge \text{SM}[\Xi;\ \emptyset]$$

Similarly, applying Theorem 13 repeatedly, we can show that the above formula is equivalent to (c).

*Between (c) and (d):* By Proposition 1. □

### C.5 Proof of Theorem 3

**Between** $(a)$ **and** $(b)$**:** Since $\mathcal{D}_{caused}$ is canonical relative to *Caused*, by Theorem 1, (a) is equivalent to

$$\mathrm{SM}[\mathcal{D}_{caused}; Caused] \wedge \mathcal{D}_{poss} \wedge \mathcal{D}^{-}_{rest} \wedge (22). \tag{41}$$

Consequently, it is sufficient to prove the claim that, under the assumption $\forall s\, Sit(s)$, formula (22) is equivalent to $\mathrm{SM}[\mathcal{D}_{sit}; Sit]$.

First note that under the assumption, (22) can be equivalently rewritten as

$$\forall p\big(p(S_0) \wedge \forall a, s(p(s) \rightarrow p(do(a,s))) \rightarrow p = Sit\big). \tag{42}$$

On the other hand, under $\forall s\, Sit(s)$, $\mathrm{SM}[\mathcal{D}_{sit}; Sit]$ is equivalent to

$$\begin{aligned}&Sit(S_0) \wedge \forall a, s(Sit(s) \rightarrow Sit(do(a,s)))\\ &\wedge\ \forall p\big(p < Sit \rightarrow \neg(p(S_0) \wedge \forall a, s(p(s) \rightarrow p(do(a,s))) \wedge \forall a, s(Sit(s) \rightarrow Sit(do(a,s))))\big),\end{aligned}$$

which, under the assumption $\forall s\, Sit(s)$, is equivalent to

$$\forall p\big(p(S_0) \wedge \forall a, s(p(s) \rightarrow p(do(a,s))) \rightarrow \neg(p < Sit)\big)$$

and furthermore to (42).

**Between** $(b)$ **and** $(c)$**:** Since $\phi(s)$ does not contain *Poss*, the equivalence follows from the equivalence between completion and the stable model semantics.

**Between** $(c)$ **and** $(d)$**:** Since $\mathcal{D}_{caused}$ contains no strictly positive occurrence of *Poss* and $\mathcal{D}_{poss^{\rightarrow}}$ contains no occurrence of *Caused*, every strongly connected component in the predicate dependency graph of $\mathcal{D}_{caused} \wedge \mathcal{D}_{poss^{\rightarrow}}$ relative to $\{Caused, Poss\}$ either belongs to $\{Caused\}$ or belongs to $\{Poss\}$. By Theorem 13, it follows that $(b)$ is equivalent to

$$\mathrm{SM}[\mathcal{D}_{caused} \wedge \mathcal{D}_{poss^{\rightarrow}}; Caused, Poss] \wedge \mathcal{D}^{-}_{rest} \wedge \mathrm{SM}[\mathcal{D}_{sit}; Sit].$$

Similarly, applying Theorem 13 two more times, we get that the above formula is equivalent to $(c)$. □

### C.6 Proof of Theorem 4

Theory $T$ is

$$\Sigma \wedge \mathcal{D}_{effect} \wedge \mathcal{D}_{precond} \wedge \mathcal{D}_{S_0} \wedge \mathcal{D}_{una} \wedge \mathcal{D}_{inertia} \wedge \mathcal{D}_{exogenous_0},$$

and the corresponding BAT is

$$\Sigma \wedge \mathcal{D}_{ss} \wedge \mathcal{D}_{ap} \wedge \mathcal{D}_{S_0} \wedge \mathcal{D}_{una}.$$

Without loss of generality, we assume that $T$ is already equivalently rewritten so that there are exactly one positive effect axiom and exactly one negative effect axiom for each fluent $R$, and that there is exactly one action precondition axiom for each action $A$.

Consider

$$\text{SM}[\Sigma \wedge \mathcal{D}_{\mathit{effect}} \wedge \mathcal{D}_{\mathit{precond}} \wedge \mathcal{D}_{S_0} \wedge \mathcal{D}_{\mathit{una}} \wedge \mathcal{D}_{\mathit{inertia}} \wedge \mathcal{D}_{\mathit{exogenous}_0};\ \mathit{Poss}, \mathit{Holds}, {\sim}\mathit{Holds}].$$

Since $\Sigma$ and $\mathcal{D}_{\mathit{una}}$ are negative on the intensional predicates, the formula is equivalent to

$$\text{SM}[\mathcal{D}_{\mathit{effect}} \wedge \mathcal{D}_{\mathit{precond}} \wedge \mathcal{D}_{S_0} \wedge \mathcal{D}_{\mathit{inertia}} \wedge \mathcal{D}_{\mathit{exogenous}_0};\ \mathit{Poss}, \mathit{Holds}, {\sim}\mathit{Holds}] \wedge \Sigma \wedge \mathcal{D}_{\mathit{una}}. \tag{43}$$

Since $\mathit{Poss}$ does not occur in

$$\mathcal{D}_{\mathit{effect}} \wedge \mathcal{D}_{S_0} \wedge \mathcal{D}_{\mathit{inertia}} \wedge \mathcal{D}_{\mathit{exogenous}_0},$$

and since $\mathcal{D}_{\mathit{precond}}$ is negative on $\{\mathit{Holds}, {\sim}\mathit{Holds}\}$, by Theorem 13, (43) is equivalent to

$$\begin{aligned}&\text{SM}[\mathcal{D}_{\mathit{effect}} \wedge \mathcal{D}_{S_0} \wedge \mathcal{D}_{\mathit{inertia}} \wedge \mathcal{D}_{\mathit{exogenous}_0};\ \mathit{Holds}, {\sim}\mathit{Holds}]\\ &\wedge \text{SM}[\mathcal{D}_{\mathit{precond}}; \mathit{Poss}] \wedge \Sigma \wedge \mathcal{D}_{\mathit{una}},\end{aligned} \tag{44}$$

which is equivalent to

$$\begin{aligned}&\text{SM}[\mathcal{D}_{\mathit{effect}} \wedge \mathcal{D}_{S_0} \wedge \mathcal{D}_{\mathit{inertia}} \wedge \mathcal{D}_{\mathit{exogenous}_0};\ \mathit{Holds}, {\sim}\mathit{Holds}]\\ &\wedge \mathcal{D}_{\mathit{ap}} \wedge \Sigma \wedge \mathcal{D}_{\mathit{una}}.\end{aligned}$$

Therefore the statement of the theorem can be proven by showing the following: if

$$I \models \neg\exists \mathbf{x}\, a\, s(\Gamma^+_R(\mathbf{x}, \mathrm{a}, \mathrm{s}) \wedge \Gamma^-_R(\mathbf{x}, \mathrm{a}, \mathrm{s})) \tag{45}$$

for every fluent $R$, and

$$I \models \Sigma \tag{46}$$

then $I$ satisfies

$$\text{SM}[\mathcal{D}_{S_0} \wedge \mathcal{D}_{\mathit{exogenous}_0} \wedge \mathcal{D}_{\mathit{effect}} \wedge \mathcal{D}_{\mathit{inertia}};\ \mathit{Holds}, {\sim}\mathit{Holds}] \tag{47}$$

iff $I|_\sigma$ satisfies

$$\mathcal{D}_{S_0} \wedge \mathcal{D}_{ss}.$$

From $\mathcal{D}_{\mathit{exogenous}_0}$, it follows that (47) is equivalent to

$$\text{SM}[\mathcal{D}^{\neg\neg}_{S_0} \wedge \mathcal{D}_{\mathit{exogenous}_0} \wedge \mathcal{D}_{\mathit{effect}} \wedge \mathcal{D}_{\mathit{inertia}};\ \mathit{Holds}, {\sim}\mathit{Holds}], \tag{48}$$

where $\mathcal{D}^{\neg\neg}_{S_0}$ is the formula obtained from $\mathcal{D}_{S_0}$ by prepending $\neg\neg$ to all occurrences of *Holds*. Under the assumption (46),

$$\mathcal{D}^{\neg\neg}_{S_0} \wedge \mathcal{D}_{\mathit{exogenous}_0} \wedge \mathcal{D}_{\mathit{effect}} \wedge \mathcal{D}_{\mathit{inertia}}$$

is $\{\mathit{Holds}\}$*-atomic-tight* w.r.t. $I$, [26] so that by the relationship between completion and SM that is stated in Corollary 11 of (Lee & Meng, 2011), we have that $I \models (48)$ iff $I$ satisfies $\mathcal{D}_{S_0}$, and, for each fluent $R$,

26. See Section 7 from the work of Lee and Meng (2011) for the definition.

$$Holds(R(\mathrm{x}), do(\mathrm{a}, \mathrm{s})) \leftrightarrow \Gamma_R^+(\mathrm{x}, \mathrm{a}, \mathrm{s}) \vee (Holds(R(\mathrm{x}, \mathrm{s}) \wedge \neg \sim Holds(R(\mathrm{x}), do(\mathrm{a}, \mathrm{s}))) \quad (49)$$

and

$$\sim Holds(R(\mathrm{x}), do(\mathrm{a}, \mathrm{s})) \leftrightarrow \Gamma_R^-(\mathrm{x}, \mathrm{a}, \mathrm{s}) \vee (\sim Holds(R(\mathrm{x}), \mathrm{s}) \wedge \neg Holds(R(\mathrm{x}), do(\mathrm{a}, \mathrm{s}))), \quad (50)$$

where x, a, s are any (lists of) object names of corresponding sorts.

It remains to show that, under the assumption (45), $I$ satisfies (49) $\wedge$ (50) iff $I|_\sigma$ satisfies

$$Holds(R(\mathrm{x}), do(\mathrm{a}, \mathrm{s})) \;\leftrightarrow\; \Gamma_R^+(\mathrm{x}, \mathrm{a}, \mathrm{s}) \vee (Holds(R(\mathrm{x}), \mathrm{s}) \wedge \neg\Gamma_R^-(\mathrm{x}, \mathrm{a}, \mathrm{s})). \quad (51)$$

In the following we will use the following facts.

- $I \models \sim Holds(R(\mathrm{x}), \mathrm{s})$ iff $I|_\sigma \not\models Holds(R(\mathrm{x}), \mathrm{s})$.
- if $F$ is a ground formula that does not contain $\sim$, then $I \models F$ iff $I|_\sigma \models F$.

*Left to Right*: Assume $I \models (49) \wedge (50)$.

- *Case 1:* $I|_\sigma \models Holds(R(\mathrm{x}), do(\mathrm{a}, \mathrm{s}))$. Clearly, $I \models Holds(R(\mathrm{x}), do(\mathrm{a}, \mathrm{s}))$, so that, from (49), there are two subcases to consider.
  - *Subcase 1:* $I \models \Gamma_R^+(\mathrm{x}, \mathrm{a}, \mathrm{s})$. Clearly, $I|_\sigma$ satisfies both LHS and RHS of (51).
  - *Subcase 2:* $I \models Holds(R(\mathrm{x}), \mathrm{s})$. From (50), it follows that $I \not\models \Gamma_R^-(\mathrm{x}, \mathrm{a}, \mathrm{s})$, and consequently, $I|_\sigma \not\models \Gamma_R^-(\mathrm{x}, \mathrm{a}, \mathrm{s})$. Clearly, $I|_\sigma$ satisfies both LHS and RHS of (51).
- *Case 2:* $I|_\sigma \not\models Holds(R(\mathrm{x}), do(\mathrm{a}, \mathrm{s}))$. It follows from (49) that $I \not\models \Gamma_R^+(\mathrm{x}, \mathrm{a}, \mathrm{s})$, which is equivalent to saying that $I|_\sigma \not\models \Gamma_R^+(\mathrm{x}, \mathrm{a}, \mathrm{s})$. Also since $I \models \sim Holds(R(\mathrm{x}), do(\mathrm{a}, \mathrm{s}))$, from (50), there are two subcases to consider.
  - *Subcase 1:* $I \models \Gamma_R^-(\mathrm{x}, \mathrm{a}, \mathrm{s})$. Clearly, $I|_\sigma$ satisfies neither LHS nor RHS of (51).
  - *Subcase 2:* $I \models \sim Holds(R(\mathrm{x}), \mathrm{s})$. This is equivalent to saying that $I|_\sigma \not\models Holds(R(\mathrm{x}), \mathrm{s})$. Clearly, $I|_\sigma$ satisfies neither LHS nor RHS of (51).

*Right to Left*: Assume $I|_\sigma \models (51)$.

- *Case 1:* $I \models Holds(R(\mathrm{x}), do(\mathrm{a}, \mathrm{s}))$. It follows from (51) that $I|_\sigma$ satisfies RHS of (51), so that there are two subcases to consider.
  - *Subcase 1:* $I|_\sigma \models \Gamma_R^+(\mathrm{x}, \mathrm{a}, \mathrm{s})$. Clearly, $I$ satisfies both LHS and RHS of (49). Also from (45), it follows that $I \not\models \Gamma_R^-(\mathrm{x}, \mathrm{a}, \mathrm{s})$. Consequently, $I$ satisfies neither LHS nor RHS of (50).
  - *Subcase 2:* $I|_\sigma \models Holds(R(\mathrm{x}), \mathrm{s}) \wedge \neg\Gamma_R^-(\mathrm{x}, \mathrm{a}, \mathrm{s})$. Clearly, $I$ satisfies both LHS and RHS of (49). Since $I \not\models \Gamma_R^-(\mathrm{x}, \mathrm{a}, \mathrm{s})$, $I$ satisfies neither LHS nor RHS of (50).
- *Case 2:* $I \models \sim Holds(R(\mathrm{x}), do(\mathrm{a}, \mathrm{s}))$. It follows from (51) that $I|_\sigma \not\models \Gamma_R^+(\mathrm{x}, \mathrm{a}, \mathrm{s})$, and $I|_\sigma \not\models (Holds(R(\mathrm{x}), \mathrm{s}) \wedge \neg\Gamma_R^-(\mathrm{x}, \mathrm{a}, \mathrm{s}))$. From the latter, consider the two subcases.

– *Subcase 1:* $I|_\sigma \not\models Holds(R(\mathrm{x}), \mathrm{s})$. Clearly, $I$ satisfies neither LHS nor RHS of (49), and satisfies both LHS and RHS of (50).

– *Subcase 2:* $I|_\sigma \not\models \neg\Gamma_R^-(\mathrm{x}, \mathrm{a}, \mathrm{s})$. Clearly, $I$ satisfies neither LHS nor RHS of (49), and satisfies both LHS and RHS of (50).

□

## C.7 Proof of Proposition 2

**Lemma 5** *Let $F$ be a formula, let $\mathbf{p}$ be a list of distinct predicate constants, let $G$ be a subformula of $F$ and let $G'$ be any formula that is classically equivalent to $G$. Let $F'$ be the formula obtained from $F$ by substituting $G'$ for $G$. If the occurrence of $G$ is in a subformula of $F$ that is negative on $\mathbf{p}$ and the occurrence of $G'$ is in a subformula of $F'$ that is negative on $\mathbf{p}$, then*

$$\mathrm{SM}[F; \mathbf{p}] \leftrightarrow \mathrm{SM}[F'; \mathbf{p}]$$

*is logically valid.*

**Proof**. Let $F^{\neg\neg}$ be the formula obtained from $F$ by prepending $\neg\neg$ to $G$, and let $(F')^{\neg\neg}$ be the formula obtained from $F'$ by prepending $\neg\neg$ to $G'$. By the Theorem on Double Negations (Theorem 14), the following formulas are logically valid.

$$\mathrm{SM}[F; \mathbf{p}] \leftrightarrow \mathrm{SM}[F^{\neg\neg}; \mathbf{p}],$$
$$\mathrm{SM}[F'; \mathbf{p}] \leftrightarrow \mathrm{SM}[(F')^{\neg\neg}; \mathbf{p}].$$

From Lemma 1, it follows that

$$(\mathbf{u} \le \mathbf{p} \wedge (G \leftrightarrow G')) \rightarrow ((F^{\neg\neg})^*(\mathbf{u}) \leftrightarrow ((F')^{\neg\neg})^*(\mathbf{u}))$$

is logically valid, where $\mathbf{u}$ is a list of predicate variables corresponding to $\mathbf{p}$. Consequently,

$$\mathrm{SM}[F^{\neg\neg}; \mathbf{p}] \leftrightarrow \mathrm{SM}[(F')^{\neg\neg}; \mathbf{p}]$$

is logically valid. □

**Proof of Proposition 2.** In formula

$$\mathrm{SM}[F' \wedge \forall \mathbf{x} y(G(y, \mathbf{x}) \rightarrow q(\mathbf{x})); \mathbf{p}, q], \tag{52}$$

clearly, $F'$ is negative on $q$ and $\forall \mathbf{x} y(G(y, \mathbf{x}) \rightarrow q(\mathbf{x}))$ is negative on $\mathbf{p}$. Let $H$ be any subformula of $F$ that is negative on $\mathbf{p}$ and contains the occurrence of $\exists y G(y, \mathbf{x})$. Consider two cases.

- Case 1: the occurrence of $\exists y G(y, \mathbf{x})$ in $H$ is not strictly positive. Thus the dependency graph of $F' \wedge \forall \mathbf{x} y(G(y, \mathbf{x}) \rightarrow q(\mathbf{x}))$ relative to $\{\mathbf{p}, q\}$ has no incoming edges into $q$.
- Case 2: the occurrence of $\exists y G(y, \mathbf{x})$ in $H$ is strictly positive. Since $H$ is negative on $\mathbf{p}$, $\exists y G(y, \mathbf{x})$ is negative on $\mathbf{p}$ as well, so that the dependency graph of $F' \wedge \forall \mathbf{x} y(G(y, \mathbf{x}) \rightarrow q(\mathbf{x}))$ relative to $\{\mathbf{p}, q\}$ has no outgoing edges from $q$.

Therefore, every strongly connected component in the dependency graph belongs to either $\mathbf{p}$ or $\{q\}$. Consequently, by Theorem 13, (52) is equivalent to

$$\mathrm{SM}[F';\mathbf{p}] \wedge \mathrm{SM}[\forall \mathbf{x} y (G(y,\mathbf{x}) \rightarrow q(\mathbf{x})); q] \tag{53}$$

Since $\exists y G(y,\mathbf{x})$ is negative on $q$, formula $\forall \mathbf{x} y (G(y,\mathbf{x}) \rightarrow q(\mathbf{x}))$ is tight on $\{q\}$. By Theorem 15, (53) is equivalent to

$$\mathrm{SM}[F';\mathbf{p}] \wedge \forall \mathbf{x}(\exists y G(y,\mathbf{x}) \leftrightarrow q(\mathbf{x})). \tag{54}$$

By Lemma 5, it follows that (54) is equivalent to

$$\mathrm{SM}[F;\mathbf{p}] \wedge \forall \mathbf{x}(\exists y G(y,\mathbf{x}) \leftrightarrow q(\mathbf{x})).$$

Consequently, the claim follows. □

### C.8 Proof of Theorem 6

It is clear that the algorithm terminates and yields a quantifier-free formula $K$. We will prove that $\mathrm{SM}[F;\mathbf{p}] \Leftrightarrow_\sigma \mathrm{SM}[\forall \mathbf{x} K; \mathbf{p} \cup \mathbf{q}]$, where $\mathbf{x}$ is the list of all (free) variables of $K$.

Let $F^{\neg\neg}$ be the formula obtained from the initial formula $F$ by prepending double negations in front of every maximal strictly positive occurrence of formulas of the form $\exists y G(\mathbf{x}, y)$. Since $F$ is almost universal relative to $\mathbf{p}$, such an occurrence is in a subformula of $F$ that is negative on $\mathbf{p}$. Thus by the Theorem on Double Negations (Theorem 14), $\mathrm{SM}[F;\mathbf{p}]$ is equivalent to $\mathrm{SM}[F^{\neg\neg};\mathbf{p}]$. Note that $F^{\neg\neg}$ contains no strictly positive occurrence of formulas of the form $\exists y G(\mathbf{x}, y)$.

For each iteration, let us assume that the formula before the iteration is

$$H_0 \wedge \cdots \wedge H_n,$$

where $H_0$ is transformed from $F^{\neg\neg}$ by the previous iterations, and each $H_i$ ($i > 0$) is a formula of the form $G(\mathbf{x}, y) \rightarrow p_G(\mathbf{x})$ that is introduced by Step (b). Initially $H_0$ is $F^{\neg\neg}$ and $n = 0$. Let $\mathbf{r}_0$ be $\mathbf{p}$, and let $\mathbf{r}_i$ be each $p_G$ for $H_i$ ($i > 0$). By induction we can prove that

(i) every positive occurrence of formulas of the form $\exists y G(\mathbf{x}, y)$ in $H_i$ is not strictly positive, and is in a subformula of $H_i$ that is negative on $\mathbf{r}_i$;

(ii) every negative occurrence of formulas of the form $\forall y G(\mathbf{x}, y)$ in $H_i$ is in a subformula of $H_i$ that is negative on $\mathbf{r}_i$.

We will prove that if Step (a) or Step (c) is applied to turn $H_k$ into $H'_k$, then

$$\mathrm{SM}[\forall \mathbf{x}_0 H_0; \mathbf{r}_0] \wedge \cdots \wedge \mathrm{SM}[\forall \mathbf{x}_n H_n; \mathbf{r}_n] \tag{55}$$

is equivalent to

$$\mathrm{SM}[\forall \mathbf{x}'_0 H'_0; \mathbf{r}_0] \wedge \cdots \wedge \mathrm{SM}[\forall \mathbf{x}'_n H'_n; \mathbf{r}_n], \tag{56}$$

where $H'_j = H_j$ for all $j$ different from $k$, and $\mathbf{x}_i$ ($i \geq 0$) is the list of all free variables of $H_i$, and $\mathbf{x}'_i$ ($i \geq 0$) is the list of all free variables of $H'_i$.

Indeed, Step (a) is a part of prenex form conversion, which preserves strong equivalence (Theorem 5). So it is clear that (55) is equivalent to (56).

When Step (c) is applied to turn (55) into (56), since $\forall y H(\mathbf{x}, y)$ is in a subformula of $H_k$ that is negative on $\mathbf{r}_k$, the equivalence between (55) and (56) follows from Lemma 5.

When Step (b) is applied to turn $H_k$ into $H'_k$ and introduces a new conjunctive term $H'_{n+1}$, formula (55) is $(\sigma, \mathbf{r}_1, \dots, \mathbf{r}_n)$-equivalent to

$$\mathrm{SM}[\forall \mathbf{x}'_0 H'_0; \mathbf{r}_0] \wedge \cdots \wedge \mathrm{SM}[\forall \mathbf{x}'_n H'_n; \mathbf{r}_n] \wedge \mathrm{SM}[\forall \mathbf{x}'_{n+1} H'_{n+1}; \mathbf{r}_{n+1}] \tag{57}$$

by Proposition 2 due to condition (i).

Let

$$H''_0 \wedge \cdots \wedge H''_m \tag{58}$$

be the final quantifier-free formula, where $H''_0$ is transformed from $F^{\neg\neg}$. By the induction, it follows that $\mathrm{SM}[F; \mathbf{p}]$ is $\sigma$-equivalent to

$$\mathrm{SM}[\forall \mathbf{x}''_0 H''_0;\ \mathbf{r}_0] \wedge \cdots \wedge \mathrm{SM}[\forall \mathbf{x}''_m H''_m;\ \mathbf{r}_m], \tag{59}$$

where each $\mathbf{x}''_i$ $(0 \le i \le m)$ is the list of all free variables of $H''_i$.

Since every non-strictly positive occurrence of new predicate $\mathbf{r}_i$ $(i > 0)$ in any $H''_j$ $(0 \le j \le m)$ is positive, there is no incoming edge into $\mathbf{r}_i$ in the dependency graph of (58) relative to $\mathbf{r}_0, \mathbf{r}_1, \dots, \mathbf{r}_m$. Consequently, every strongly connected component of the dependency graph belongs to one of $\mathbf{r}_i$ $(i \ge 0)$. Moreover, it is clear that each $H''_i$ $(i \ge 0)$ is negative on every $\mathbf{r}_j$ for $j \ne i$. (In the case of $H''_0$, recall that the occurrence of $\mathbf{r}_j$ for any $j > 0$ is not strictly positive since $F^{\neg\neg}$, from which $H''_0$ is obtained, contains no strictly positive occurrence of formulas of the form $\exists y G(\mathbf{x}, y)$.) Thus by the splitting theorem (Theorem 13), formula (59) is equivalent to

$$\mathrm{SM}[\forall \mathbf{x}''_0 H''_0 \wedge \cdots \wedge \forall \mathbf{x}''_m H''_m;\ \mathbf{r}_0 \cup \cdots \cup \mathbf{r}_m]. \tag{60}$$

□

### C.9 Proof of Theorem 7

We use the notations introduced in the proof of Theorem 6. By Theorem 6, $\mathrm{SM}[F; \mathbf{p}]$ is $\sigma$-equivalent to (60) and, by Theorem 12, (60) is equivalent to

$$\mathrm{SM}[\forall \mathbf{x}''_0 H''_0 \wedge \cdots \wedge \forall \mathbf{x}''_m H''_m \wedge \mathit{Choice}(\sigma^{pred} \setminus \mathbf{p});\ \sigma^{pred} \cup \mathbf{r}_1 \cup \cdots \cup \mathbf{r}_m] \tag{61}$$

($\mathbf{r}_0$ is $\mathbf{p}$), where $\sigma^{pred}$ is the set of all predicate constants in signature $\sigma$. It follows from Proposition 3 from (Cabalar et al., 2005) that (61) is equivalent to

$$\mathrm{SM}[\forall \mathbf{x}''_0 H'''_0 \wedge \cdots \wedge \forall \mathbf{x}''_m H'''_m \wedge \mathit{Choice}(\sigma^{pred} \setminus \mathbf{p});\ \sigma^{pred} \cup \mathbf{r}_1 \cup \cdots \cup \mathbf{r}_m], \tag{62}$$

where $H'''_i$ is obtained from $H''_i$ by applying the translation from (Cabalar et al., 2005, Section 3) that turns a quantifier-free formula into a set of rules. It is easy to see that $F'$ is the same as the formula

$$\forall \mathbf{x}''_0 H'''_0 \wedge \cdots \wedge \forall \mathbf{x}''_m H'''_m \wedge \mathit{Choice}(\sigma^{pred} \setminus \mathbf{p})$$

and $\sigma^{pred} \cup \mathbf{r}_1 \cup \cdots \cup \mathbf{r}_m$ is the same as $\mathbf{p} \cup pr(F')$, so that (62) can be written as

$$\text{SM}[F';\ \mathbf{p} \cup pr(F')],$$

which is equivalent to

$$\text{SM}[F' \wedge \mathit{False}(\mathbf{p} \setminus pr(F'))].$$

by Proposition 1. □

## C.10 Proof of Theorem 8

Assume that $T$ is

$$\begin{array}{l}\text{CIRC}[\Sigma; \mathit{Initiates}, \mathit{Terminates}, \mathit{Releases}] \wedge \text{CIRC}[\Delta; \mathit{Happens}] \\ \wedge\ \text{CIRC}[\Theta; Ab_1, \ldots, Ab_n] \wedge \Xi,\end{array}$$

which is equivalent to

$$\begin{array}{l}\text{SM}[\Sigma; \mathit{Initiates}, \mathit{Terminates}, \mathit{Releases}] \wedge \text{SM}[\Delta; \mathit{Happens}] \\ \wedge\ \text{SM}[\Theta; Ab_1, \ldots, Ab_n] \wedge \Xi\end{array} \tag{63}$$

by Theorem 2.

Let $\Xi_{def}$ be the set of all definitions (35) in $\Xi$, and let $\Xi'$ be the formula obtained from $\Xi$ by applying Step 1. By Theorem 15, it follows that each formula (35) in $\Xi_{def}$ is equivalent to

$$\text{SM}[\forall \mathbf{x}(G' \rightarrow p(\mathbf{x}));\ p],$$

where $G'$ is as described in Step 1. Consequently, (63) is equivalent to

$$\begin{array}{l}\text{SM}[\Sigma; \mathit{Initiates}, \mathit{Terminates}, \mathit{Releases}] \wedge \text{SM}[\Delta; \mathit{Happens}] \\ \wedge\ \text{SM}[\Theta; Ab_1, \ldots, Ab_n] \wedge \bigwedge_{(35) \in \Xi_{def}} \text{SM}[\forall \mathbf{x}(G' \rightarrow p(\mathbf{x}));\ p] \wedge \Xi'',\end{array} \tag{64}$$

where $\Xi''$ is the conjunction of all the axioms in $\Xi'$ other than the ones obtained from definitional axioms (35).

Applying Theorem 13 repeatedly, it follows that (64) is equivalent to

$$\begin{array}{l}\text{SM}[\Sigma \wedge \Delta \wedge \Theta \wedge \Xi'' \wedge \bigwedge_{(35) \in \Xi_{def}} \forall \mathbf{x}(G' \rightarrow p(\mathbf{x})); \\ \mathit{Initiates}, \mathit{Terminates}, \mathit{Releases}, \mathit{Happens}, Ab_1, \ldots, Ab_n, \mathbf{p}]\ .\end{array} \tag{65}$$

According to the syntax of the event calculus reviewed in Section 3.1,

- every positive occurrence of a formula of the form $\exists y G(y)$ in (65) is contained in a subformula that is negative on $\{\mathit{Initiates}, \mathit{Terminates}, \mathit{Releases}, \mathit{Happens}, Ab_1, \ldots, Ab_n, \mathbf{p}\}$, and
- there are no negative occurrences of any formula of the form $\forall y G(y)$ in (65).

Consequently, the statement of the theorem follows from Theorem 7. □

### C.11 Proof of Theorem 9

Since (37) is almost universal relative to $\{Caused, Poss, Sit\}$, the result follows from Theorems 7 and 3. □

### C.12 Proof of Theorem 10

From $\mathcal{D}_{exogenous_0}$, it follows that $\mathrm{SM}[T; Holds, {\sim}Holds, Poss]$ is equivalent to $\mathrm{SM}[T^{\neg\neg}; Holds, {\sim}Holds, Poss]$, where $T^{\neg\neg}$ is obtained from $T$ by prepending $\neg\neg$ to all occurrences of *Holds* in $\mathcal{D}_{S_0}$. From the definition of a uniform formula (Reiter, 2001), it follows that $T^{\neg\neg}$ is almost universal relative to $\{Holds, {\sim}Holds, Poss\}$. The result follows from Theorem 7. □